%% file: main.tex
\documentclass[letterpaper, 10 pt, journal, twoside]{IEEEtran}

\usepackage[T1]{fontenc}
\usepackage[english]{babel}
\usepackage[pdftex]{graphicx}   
\usepackage{epsfig}             
\usepackage{amsmath}            
\usepackage{amssymb}            
\usepackage{mathtools}
\usepackage{multicol}
\usepackage[bookmarks=true]{hyperref}
\usepackage{xcolor}
\usepackage{siunitx}
\usepackage{lipsum}
\usepackage[normalem]{ulem}
\usepackage[ruled]{algorithm2e}
\usepackage{url}

\newcommand\cst{\bgroup\markoverwith
{\textcolor{red}{\rule[0.5ex]{2pt}{1.5pt}}}\ULon}
\newcommand\cstw[1]{\bgroup\markoverwith
{\textcolor{red}{\rule[#1ex]{2pt}{1.5pt}}}\ULon}
\definecolor{mod}{RGB}{0, 153, 77}

\usepackage{bm}

\input{preamble/preamble}
\input{preamble/notation_defs}

\input{styles/math}

\input{styles/robotics}
\input{styles/references}

\usepackage{acro}
\input{acro/acro}

\usepackage[caption=false,font=footnotesize]{subfig}

\usepackage{ctable}

\newcommand{\ts}{\textsuperscript}
\DeclareSIUnit{\sqrts}{\ensuremath{\sqrt{\text{\second}}}}

\hypersetup{
  pdfinfo={
    Title={Overcoming Bias: Equivariant Filter Design for Biased Attitude Estimation with Online Calibration},
    Author={Alessandro Fornasier, Yonhon Ng, Christian Brommer, Christoph B\"ohm, Robert Mahony and Stephan Weiss},
    Subject={Equivariant Filter},
    Keywords={Equivariant Filter, Equivariant System, State Estimation, Attitude, Directions, Navigation, EQF, IEKF, Bias}
  }
}

\title{Overcoming Bias: Equivariant Filter Design for Biased Attitude Estimation with Online Calibration}

\author{Alessandro Fornasier$^{1}$, Yonhon Ng$^{2}$, Christian Brommer$^{1}$, Christoph B\"ohm$^{1}$, Robert Mahony$^{2}$ and Stephan Weiss$^{1}$%
\thanks{This work was supported by the EU-H2020 project BUGWRIGHT2 (GA 871260), and by the Australian Research Council through Discovery Grant DP210102607.}
\thanks{$^{1}$Alessandro Fornasier, Christian Brommer, Christoph B\"ohm and Stephan Weiss are with the Control of Networked Systems Group, University of Klagenfurt, Austria. {\tt\small \{name.surname\}@ieee.org}}%
\thanks{$^{2}$Yonhon Ng and Robert Mahony are with the System Theory and Robotics Lab, Australian National University, Australia. {\tt\small \{name.surname\}@anu.edu.au}}%
}

\begin{document}

\maketitle


\input{sections/abstract}


\noindent{\footnotesize\ttfamily An implementation of the presented EqF is available at: \url{https://github.com/aau-cns/ABC-EqF}.}

\input{sections/introduction}
\input{sections/notation}
\input{sections/system}
\input{sections/eqf}
\input{sections/comparison}
\input{sections/experiment}
\input{sections/conclusion}

\bibliographystyle{IEEEtran}
\bibliography{bibliography/IEEEabrv,bibliography/main,bibliography/b04_2_115,bibliography/extended}


\end{document}


\maketitle
\thispagestyle{empty}
\pagestyle{empty}


This document provides a gentle introduction to differential geometry concepts such as group actions, homogeneous spaces, and semi-direct product group, lemma and theorems proofs, as well as EqF, and IEKF design notes and algorithms. This document is intended to help the reader to better familiarize with the derivation of an Equivariant Filter, and to highlight the differences between the an EqF, and the state-of-the-art IEKF for the biased attitude estimation with online calibration problem.

\input{sections_supp/notation}
\input{sections_supp/preliminaries}
\input{sections_supp/eqf}
\input{sections_supp/iekf}
\input{sections_supp/appendix}

\bibliographystyle{IEEEtran}
\bibliography{bibliography/IEEEabrv,bibliography/main}


%% file: preamble/preamble.tex
\usepackage[T1]{fontenc}

\usepackage{MnSymbol}
\usepackage{mathdots}

\usepackage{amsthm}



%% file: preamble/notation_defs.tex
\providecommand{\R}{\mathbb{R}}

\providecommand{\SO}{\mathbf{SO}}

\providecommand{\grpG}{\mathbf{G}}

\providecommand{\gothso}{\mathfrak{so}}

\providecommand{\gothg}{\mathfrak{g}}

\providecommand{\calM}{\mathcal{M}}
\providecommand{\calN}{\mathcal{N}}

\providecommand{\torSO}{\mathcal{SO}}

\providecommand{\vecL}{\mathbb{L}}

\providecommand{\PD}{\mathbb{S}_+} 

\providecommand{\td}{\mathrm{d}}

\providecommand{\Fr}[2]{\left. \mathrm{D}_{#1} \right|_{#2}}

\usepackage{accents}
\makeatletter
\providecommand{\scirc}{%
    \hbox{\fontfamily{\rmdefault}\fontsize{0.4\dimexpr(\f@size pt)}{0}\selectfont{\raisebox{-0.52ex}[0ex][-0.52ex]{$\circ$}}}}

\makeatother

\mathchardef\mhyphen="2D



%% file: styles/math.tex
\theoremstyle{plain}
\newtheorem{theorem}{Theorem}[section]

\newtheorem{lemma}[theorem]{Lemma}

\theoremstyle{definition}

\theoremstyle{remark}
\newtheorem{remark}[theorem]{Remark}

\newcommand{\norm}[1]{\lVert #1 \rVert}

\newcommand{\AtoB}[2]{\;:\;#1\;\rightarrow\;#2}

\providecommand{\td}{\mathrm{d}}

\providecommand{\Fr}[2]{\mathrm{D}_{#1}\big|_{#2}}

\newcommand{\eye}{\mathbf{I}}



%% file: styles/robotics.tex
\newcommand{\Rot}[2]{\prescript{#1}{#2}{\mathbf{R}}}
\newcommand{\Vector}[3]{\prescript{#1}{}{\bm{#2}}_{#3}}

\newcommand{\dotRot}[2]{\prescript{#1}{#2}{\dot{\mathbf{R}}}}
\newcommand{\dotVector}[3]{\prescript{#1}{}{\dot{\bm{#2}}}_{#3}}

\newcommand{\hatVector}[3]{\prescript{#1}{}{\hat{\bm{#2}}}_{#3}}

\newcommand{\frameofref}[1]{\{$#1$\}}

%% file: styles/references.tex
\newcommand{\figref}[1]{Fig.~\ref{fig:#1}}

\newcommand{\equref}[1]{Equ.~(\ref{eq:#1})}
\newcommand{\algref}[1]{Alg.~\ref{alg:#1}}
\newcommand{\tabref}[1]{Tab.~\ref{tab:#1}}
\newcommand{\cit}[1]{~\cite{#1}}

%% file: acro/acro.tex
\DeclareAcronym{mav} {
    short   = MAV,
    long    = Micro Aerial Vehicle,
}

\DeclareAcronym{uav} {
    short   = UAV,
    long    = Unmanned Aerial Vehicle,
}

\DeclareAcronym{imu}{
    short   = IMU,
    long    = Inertial Measurement Unit,
}

\DeclareAcronym{lidar}{
    short   = LIDAR,
    long    = Laser Imaging Detection and Ranging,
}

\DeclareAcronym{vins}{
    short   = VINS,
    long    = Visual-Inertial Navigation System,
}

\DeclareAcronym{bins}{
    short   = BINS,
    long    = biased inertial navigation system,
}

\DeclareAcronym{bas}{
    short   = BAS,
    long    = biased attitude system,
}

\DeclareAcronym{vinseval}{
    short   = VINSEval,
    long    = \acs{vins} Evaluation Framework,
}

\DeclareAcronym{mcsf}{
    short   = MCSF,
    long    = Multi Copter Simulation Framework,
}

\DeclareAcronym{vio}{
    short   = VIO,
    long    = Visual-Inertial Odometry,
}

\DeclareAcronym{slam}{
    short   = SLAM,
    long    = Simultaneous Localization and Mapping,
}

\DeclareAcronym{vislam}{
    short   = VI-SLAM,
    long    = Visual-Inertial \acs{slam},
}

\DeclareAcronym{ros}{
    short   = ROS,
    long    = Robot Operating System,
}

\DeclareAcronym{nees}{
    short   = NEES,
    long    = Normalized Estimation Error Squared,
}

\DeclareAcronym{anees}{
    short   = A\acs{nees},
    long    = Average \ac{nees},
}

\DeclareAcronym{rmse}{
    short   = RMSE,
    long    = Root Mean Square Error,
}

\DeclareAcronym{ate}{
    short   = ATE,
    long    = Absolute Trajectory Error,
}

\DeclareAcronym{rte}{
    short   = RTE,
    long    = Relative Trajectory Error,
}

\DeclareAcronym{bp}{
    short   = BP,
    long    = Breaking Point,
}

\DeclareAcronym{mems}{
    short   = MEMS,
    long    = Micro Electro Mechanical Systems,
}

\DeclareAcronym{mse}{
    short   = MSE,
    long    = Mean Squared Error Matrix,
}

\DeclareAcronym{eqf}{
    short   = EqF,
    long    = Equivariant Filter,
}

\DeclareAcronym{kf}{
    short   = KF,
    long    = Kalman Filter,
}

\DeclareAcronym{ekf}{
    short   = EKF,
    long    = Extended Kalman Filter,
}

\DeclareAcronym{mekf}{
    short   = MEKF,
    long    = Multiplicative \acs{ekf},
}

\DeclareAcronym{ukf}{
    short   = UKF,
    long    = Unscented Kalman Filter,
}

\DeclareAcronym{iekf}{
    short   = IEKF,
    long    = Invariant Extended Kalman Filter,
}

\DeclareAcronym{wgn}{
    short   = WGN,
    long    = White Gaussian Noise,
}

\DeclareAcronym{gnss}{
    short   = GNSS,
    long    = Global Navigation Satellite System,
}

\DeclareAcronym{rtk}{
    short   = RTK,
    long    = Real Time Kinematics,
}

%% file: sections/abstract.tex
\begin{abstract}

Stochastic filters for on-line state estimation are a core technology for autonomous systems. 
The performance of such filters is one of the key limiting factors to a system's capability. 
Both asymptotic behavior (e.g.,~for regular operation) and transient response (e.g.,~for fast initialization and reset) of such filters are of crucial importance in guaranteeing robust operation of autonomous systems. 

This paper introduces a new generic formulation for a gyroscope aided attitude estimator using $N$ direction measurements including both body-frame and reference-frame direction type measurements.
The approach is based on an integrated state formulation that incorporates navigation, extrinsic calibration for all direction sensors, and gyroscope bias states in a single equivariant geometric structure. 
This newly proposed symmetry allows modular addition of different direction measurements and their extrinsic calibration while maintaining the ability to include bias states in the same symmetry.
The subsequently proposed filter-based estimator using this symmetry noticeably improves the transient response, and the asymptotic bias and extrinsic calibration estimation compared to state-of-the-art approaches. 
The estimator is verified in statistically representative simulations and is tested in real-world experiments.
\end{abstract}

%% file: sections/introduction.tex
\section{Introduction and Related Work}

\IEEEPARstart{A}{ttitude} estimators fuse measurements of angular velocity, obtained from strap-down gyroscopes, with partial attitude measurements, such as magnetometer sensors, horizon sensors, \ac{gnss}, etc.  
Arguably the most widely used attitude filter is the \ac{mekf}\cit{markley2003attitude}. It exploits the geometric structure  of the special orthogonal group of rotation matrices (or the locally isomorphic quaternion double cover of this group) and builds an error state Kalman filter for an intrinsic state-error linearised at the origin. 
That said, to the authors' best understanding, the state-of-the-art attitude filter presently available in the robotics community is based on the recently proposed \ac{iekf}, by Barrau and Bonnabel\cit{7523335}. 
This filter is a generalization of the left-invariant Kalman filter\cit{bonnabel2007left} which, when applied to the special orthogonal group for attitude estimation\cit{Barrau2015IntrinsicEstimation}, specializes to the \ac{mekf}. 
Although the modern filters specialise to the established \ac{mekf} when applied to the attitude estimation problem, the more recent work in\cit{7523335} provides theoretical justification, through the understanding of the ``group affine'' property, for the performance of this filter architecture. 

\begin{figure}[!t]
\centering
\includegraphics[width=0.925\linewidth]{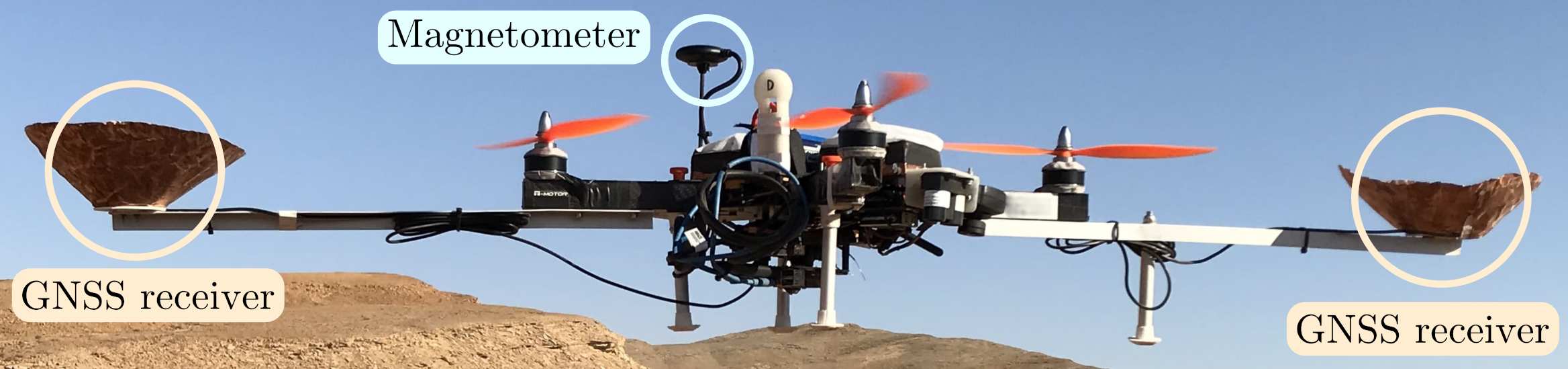}

\bigskip
\includegraphics[width=0.95\linewidth]{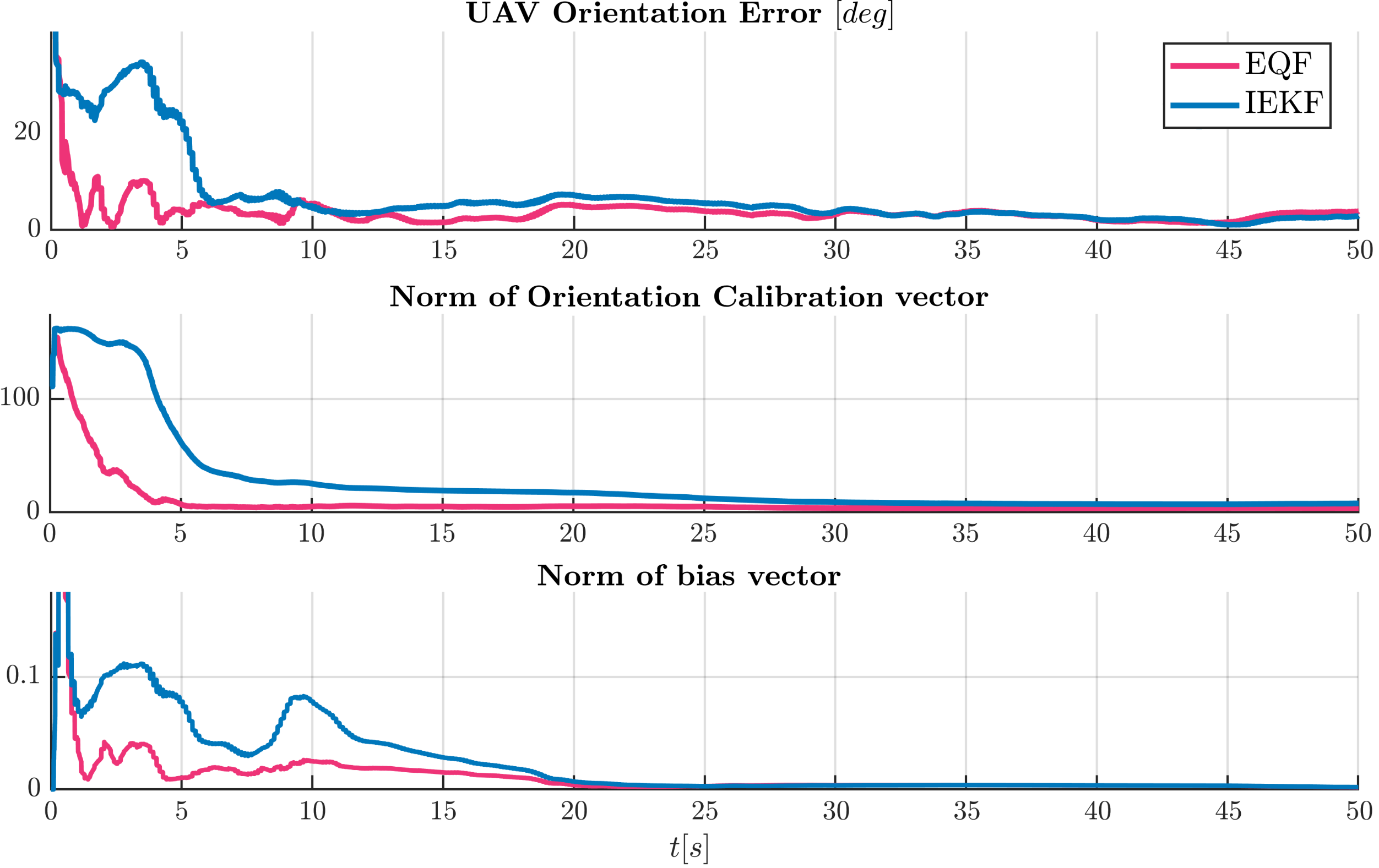}
\caption{Top (picture): Platform used for the outdoor experiments with two \acs{rtk} \acs{gnss} receivers and a magnetometer (integrated within the compass module).
Bottom (plots): Transient phase (first \SI{50}{\second} after noticeably wrong initialization) showing the norm of the errors of the estimated attitude, angular calibration vector between magnetometer and body frame, and angular velocity bias. Our proposed \ac{eqf} shows noticeably better performance compared to the state-of-the-art \ac{iekf} (cf. Section \ref{sec:exp_outdoor} for details).
}
\label{fig:twins_plots}
\vspace{-15pt}
\end{figure}

A key limitation in said prior work is the separation of sensor bias states and calibration states from the navigation states. 
In particular, although the symmetry of the navigation states (attitude) are carefully modeled in the MEKF and \ac{iekf} designs, the bias is modeled as an element of a linear space "tacked on" to the linearisation of the navigation states. 
A key observation \cite{barrau:tel-01247723} is that the group-affine property only holds for the model of the navigation states and is lost for the combined system when bias is added in this manner. 
The practical applicability of this approach, named ``Imperfect-\ac{iekf}'' by Barrau\cit{barrau:tel-01247723}, has been confirmed by several authors \cit{barrau:tel-01247723},\cit{doi:10.1177/0278364919894385},\cit{Cohen2020NavigationEstimation}.
Recent work\cit{Ng2020EquivariantGroups} on second order systems on Lie-groups has demonstrated that there is a natural symmetry that can be used to model both configuration and velocity states in a single geometry. 
This symmetry draws from advances in the field of equivariant system theory and observer design\cit{Mahony2021EquivariantGroups},\cit{Mahony2020EquivariantDesign},\cit{Ng2020EquivariantGroups},\cit{VanGoor2020EquivariantSpaces} and provides a framework for stochastic filter design that applies to a broader class of systems than the \ac{iekf} while specialising back to the  \ac{iekf} for group-affine systems on Lie-groups\cit{vanGoor2020EquivariantEqF}. In\cit{fornasier22eqf}, a new symmetry was introduced to couple the navigation state as well as the accelerometer and gyroscope biases in a single geometric structure. This was only possible by extending the system state with an artificial velocity bias state. No calibration states were included in that symmetry. 
Apart of this non-minimal state vector and very system-specific symmetry, another shortcoming of prior work, including\cit{van2020constructive},\cit{Mahony2020EquivariantWild},\cit{Goor2021AnOdometry}, is the missing discussion about an efficient integration scheme in the actual estimator implementation for integrating the continuous time filter equations derived from the found symmetry. 

In this paper, we tackle shortcomings of current state of the art through the following contributions:

{\bf (i):} We propose a new integrated equivariant symmetry; a group structure of ${\left(\SO\left(3\right) \ltimes \gothso\left(3\right)\right) \times \SO(3)^n}$ with $n \leq N$ calibration states for the generalized attitude estimation problem (inspired from\cit{fornasier22eqf}) that includes biases in a minimal state vector and which is modularly extensible to $N$ sensors with their calibration states measuring both body-frame or spatial reference directions.
    
{\bf (ii):} We propose a computationally efficient discrete-time \ac{eqf} realization based on the found symmetry so we can handle real-world data including multi-rate sensors with measurement dropouts, calibration states for body-frame and spatial direction measurements from $N$ sensors, and without hardware time synchronization. 
We leverage a Lie group integrator scheme and the derivation of a closed-form state transition matrix for discrete covariance propagation in our \ac{eqf} formulation.
    
{\bf (iii):} We perform a statistically relevant comparison against state of the art in simulation and show on both real and simulated data significantly improved performance reducing the transient time for all states and improving asymptotic behavior particularly for the bias states.

A gentle introduction on differential geometry concepts such as group actions, homogeneous spaces, and semi-direct product group, as well as all mathematical proofs, and our here-mentioned reference-implementation of an Imperfect-\ac{iekf} estimator with bias and calibration states can be found in our report\cit{supp_seafile}. 

Our proposed symmetry covers the majority of real-world partial attitude measurements such as body-frame direction measurements with known spatial reference (magnetometer, sun sensor, or horizon measurements, \emph{etc}) and spatial direction measurements with known body-frame reference (multi-\ac{gnss} measurements, \ac{gnss} velocity for nonholonomic vehicles, \emph{etc}).
The flexibility, performance and robustness of the proposed filter makes it a natural choice for real-world systems that require high quality attitude, sensor extrinsic and bias calibration with unknown initial values, and rapid filter convergence allowing in-flight resets and extrinsic calibration changes.

%% file: sections/notation.tex
\section{Notation and Preliminaries}

In this paper we use the following notation.
Let \frameofref{A} and \frameofref{B} denote different frames of reference.
In general, vectors describing physical quantities expressed in a frame of reference \frameofref{A} are denoted by $\Vector{A}{x}{}$. The rotation matrix $\Rot{A}{B}$ transforms a vector $\Vector{B}{x}{}$ written in coordinates of \frameofref{B} into a vector ${\Vector{A}{x}{} = \Rot{A}{B}\Vector{B}{x}{}}$ written in coordinates of  \frameofref{A}.

%% file: sections/system.tex
\section{Biased Attitude System}\label{sec:simp_sys}

\subsection{System Kinematics}
Let \frameofref{G} denote the global inertial frame of reference, \frameofref{I} denote the gyroscope frame of reference, and \frameofref{S_i} denote the frame of reference of the $i\,$\ts{th} sensor providing direction measurements. 
In this letter we focus on the problem of estimating the rigid body orientation ${\Rot{G}{I}}$ of a moving rigid platform, as well as the gyroscope bias ${\Vector{I}{b_{\bm{\omega}}}{}}$, and the extrinsic calibration ${\Rot{I}{S_i}}$ of the $i\,$\ts{th} direction sensors.
In non-rotating, flat earth assumption the deterministic (noise-free) system takes the following general form
\begin{subequations}\label{eq:orig_bas}
    \begin{align}
        &\dotRot{G}{I} = \Rot{G}{I}\left(\Vector{I}{\bm{\bm{\omega}}}{} - \Vector{I}{b_{\bm{\omega}}}{}\right)^{\wedge} ,\\
        &\dotVector{I}{b_{\bm{\omega}}}{} = \mathbf{0} ,\\
        &\dotRot{I}{S_i} = \mathbf{0}^{\wedge} \quad \forall\;i=1,\dots,n.
    \end{align}
\end{subequations}
where $\Vector{I}{\bm{\omega}}{}$ is the body-fixed, biased angular velocity measurements provided by the gyroscope. For the sake of generality, note that $n \leq N$, where $N$ is the total number of direction sensors, since there could exist sensors that are already calibrated.

Let ${\gamma = \left(\Rot{G}{I},\, \Vector{I}{b}{\bm{\omega}}\right) \in \torSO\left(3\right) \times \R^{3}}$ denote the system core state, and define ${\bm{\mu} = \left(\Vector{I}{\bm{\omega}}{},\, \mathbf{0}\right) \subseteq \R^6}$. Let ${\zeta = \left(\Rot{I}{S_1},\, \dots,\, \Rot{I}{S_n}\right) \in \torSO\left(3\right)^n}$ denote the sensors extrinsic calibration states, and define ${\bm{\upsilon} = \left(\mathbf{0},\, \dots,\, \mathbf{0}\right) \subseteq \R^{3n}}$. Then, the full state of the system writes $\xi = {\left(\gamma,\,\zeta\right) \in \calM \coloneqq \torSO\left(3\right) \times \R^{3} \times \torSO\left(3\right)^n}$, while the system input writes ${\mathbf{u} = \left(\bm{\mu},\,\bm{\upsilon}\right) \in \mathbb{L} \subseteq \R^{6+3n}}$.
The full system kinematics in \equref{orig_bas} can be written in the compact form ${\dot{\xi} = f_0\left(\xi\right) + f_{\mathbf{u}}\left(\xi\right)}$ as follows
\begin{equation}\label{eq:bas}
    \dot{\xi} = \left(-\Rot{G}{I}\Vector{I}{b}{\bm{\omega}}^{\wedge},\,\mathbf{0}^{\wedge},\,\dots,\,\mathbf{0}^{\wedge}\right) + \left(\Rot{G}{I}\Vector{I}{\bm{\omega}}{}^{\wedge},\,\mathbf{0}^{\wedge},\,\dots,\,\mathbf{0}^{\wedge}\right) ,
\end{equation}

For the estimation problem we consider the measurements of $N$ known directions ${\Vector{G}{d}{1},\, \dots,\, \Vector{G}{d}{N}}$, to be available to the system. The output space is then defined to be $\calN \coloneqq \R^{3N}$, and therefore, the configuration output $h \AtoB{\calM}{\calN}$ is written 
\begin{equation}\label{eq:confout_bas}
    \begin{split}
        h\left(\xi\right) = &\left(\Rot{I}{S_1}^T\Rot{G}{I}^T\Vector{G}{d}{1},\,\dots\,\Rot{I}{S_n}^T\Rot{G}{I}^T\Vector{G}{d}{n},\,\dots\,,\right.\\
        &\left.\Rot{G}{I}^T\Vector{G}{d}{n+1},\,\dots\,,\Rot{G}{I}^T\Vector{G}{d}{N}\right) \in \calN .
    \end{split}
\end{equation}

\subsection{Equivariance of the Biased Attitude System}

For the sake of clarity, in the two following sections, we omit all the superscript and subscript associated with reference frames from the state variables and input to improve the readability of the definitions and theorems. 
Moreover, without loss of generality, we restrict our analysis to the case of $N=2$ direction sensors, one with a related calibration state and one assumed fix calibrated, thus $n=1$. This covers the general case for the filter derivation since the following theorems hold for every $n,\,N>0$. 
Any other case with multiple sensors and multiple calibration states can be easily derived from the presented derivation in the following sections.
Note that we carried out a nonlinear observability analysis\cit{StephanM.Weiss2012VisionHelicopters} confirming that $N = 2$ non-parallel measurements is a sufficient condition to render the system fully observable even in the presence of no motion. 

Therefore, we consider the full system state ${\xi = \left(\left(\Rot{}{},\, \Vector{}{b}{}\right),\, \mathbf{C}\right) \in \calM}$, where $\Rot{}{}$ represents the attitude, ${\Vector{}{b}{}}$ the gyroscope bias, and ${\mathbf{C}}$ the calibration state. 
We also consider the output map ${h\left(\xi\right) = \left(\mathbf{C}^T\Rot{}{}^T\Vector{}{d}{1},\,\Rot{}{}^T\Vector{}{d}{2}\right)} \in \calN$.

Before diving directly into the derivation of the proposed \ac{eqf} for the \acl{bas}, let us quickly recall the steps involved in such derivation in order to provide a quick overview to the reader\cit{Mahony2020EquivariantDesign},\cit{VanGoor2020EquivariantSpaces},\cit{vanGoor2020EquivariantEqF}.
Our first major contribution is to define a new adequate symmetry group $\grpG$ for the defined \acl{bas}. We propose an action $\phi$ of the symmetry group on the state space $\calM$, a second action $\psi$ of the symmetry group on the input space $\mathbb{L}$ of the system, and a third action $\rho$ of the symmetry group on the output space $\calN$ of the system. 
We then prove that the system, and the output, are equivariant under our newly defined actions. 
Finally, we define a new geometric structure $\Lambda$ called lift, mapping inputs of the original system to inputs on the Lie algebra of the symmetry group.
The equivariance of the system, together with the lift, allow us to construct a new system (the lifted system) on the symmetry group and to design a filter for such a system, the proposed \ac{eqf}.

Let $X = \left(\left(A,\,a\right),\,B\right)$ be an element of the symmetry group $\grpG \coloneqq \left(\SO\left(3\right) \ltimes \gothso\left(3\right)\right) \times \SO\left(3\right)$.

\begin{lemma}
Define ${\phi \AtoB{\grpG \times \calM}{\calM}}$ as
\begin{equation}\label{eq:bas_phi}
    \phi\left(X, \xi\right) \coloneqq \left(\Rot{}{}A,\, A^T\left(\Vector{}{b}{} - a^{\vee}\right),\, A^T\mathbf{C}B\right) \in \calM .
\end{equation}
Then, $\phi$ is a transitive right group action of $\grpG$ on $\calM$.
\end{lemma}

\begin{remark}
Note that in case of multiple calibration states, where $n>1$, we would have ${X = \left(\left(A,\,a,\right)\,B,\,C\,\dots\right)}$ and ${\phi\left(X, \xi\right) \coloneqq \left(\left(\Rot{}{}A,\, A^T\left(\Vector{}{b}{} - a^{\vee}\right)\right),\, A^T\mathbf{C}_1 B,\, A^T\mathbf{C}_2 C,\,\dots\right)}$ 
\end{remark}

\begin{lemma}
Define ${\psi \AtoB{\grpG \times \vecL}{\vecL}}$ as
\begin{equation}\label{eq:bas_psi}
    \begin{split}
        \psi\left(X,\mathbf{u}\right) &\coloneqq \left(A^T\left(\bm{\omega} - a^{\vee}\right),\, \mathbf{0},\, \mathbf{0}\right) ,
    \end{split}
\end{equation}
Then, $\psi$ is a right group action of $\grpG$ on $\vecL$.
\end{lemma}

\begin{remark}
Note that in case of multiple calibration states, where $n>1$, we would simply repeat the last zero entry $n$ time. Therefore, we would have ${\psi\left(X,\mathbf{u}\right) \coloneqq \left(A^T\left(\bm{\omega} - a^{\vee}\right),\, \mathbf{0},\, \mathbf{0},\, \mathbf{0},\, \dots\right)}$
\end{remark}

\begin{theorem}
The \acl{bas} in \equref{bas} is equivariant under the actions $\phi$ in \equref{bas_phi} and $\psi$ in \equref{bas_psi} of the symmetry group $\grpG$. That is
\begin{equation*}
    f_0\left(\xi\right) + f_{\psi_{X}\left(\bm{u}\right)}\left(\xi\right) = \Phi_{X}f_0\left(\xi\right) + \Phi_{X}f_{\bm{u}}\left(\xi\right) .
\end{equation*}
\end{theorem}

\subsection{Equivariance of the Output}
\begin{lemma}
Define ${\rho \AtoB{\grpG \times \calN}{\calN}}$ as
\begin{equation}\label{eq:rho}
    \rho\left(X,\Vector{}{y}{}\right) \coloneqq \left(B^T\Vector{}{y}{1},\,A^T\Vector{}{y}{2}\right) .
\end{equation}
Then, the configuration output defined in \equref{confout_bas} is equivariant\cit{VanGoor2020EquivariantSpaces}.
\end{lemma}

\begin{remark}
Here is where we have the major changes when explicitly considering the extrinsic calibration of a sensor. 
In particular, note that for any sensor that does not have an associated calibration states the ${\rho}$ action is defined by ${A^T\Vector{}{y}{}}$, where ${\Vector{}{y}{}}$ is the sensor measurement, whereas for any sensor that does have an associated calibration states, the ${\rho}$ action is the sensor measurement ${\Vector{}{y}{}}$ pre-multiplied by the element of the symmetry group relative to the sensor extrinsic calibration. 
More concretely, in the case of ${N=n=2}$ the ${\rho}$ action is ${\rho\left(X,\Vector{}{y}{}\right) \coloneqq \left(B^T\Vector{}{y}{1},\,C^T\Vector{}{y}{2}\right)}$, in the case of ${N=2, n=0}$ the ${\rho}$ action is ${\rho\left(X,\Vector{}{y}{}\right) \coloneqq \left(A^T\Vector{}{y}{1},\,A^T\Vector{}{y}{2}\right)}$, in the mixed case the ${\rho}$ action is as shown in \equref{rho}.
\end{remark}

\subsection{Equivariant Lift and Lifted System}

\begin{theorem}
Define ${\Lambda \AtoB{\calM \times \vecL} \gothg}$ as
\begin{equation}\label{eq:lift_bas}
    \Lambda\left(\xi, \mathbf{u}\right) \coloneqq \left(\left(\left(\Vector{}{\bm{\omega}}{}^{\wedge} - \Vector{}{b}{}^{\wedge}\right),\, - \Vector{}{\omega}{}^{\wedge}\Vector{}{b}{}\right),\, \mathbf{C}^T\left(\Vector{}{\bm{\omega}}{}^{\wedge} - \Vector{}{b}{}^{\wedge}\right)\mathbf{C}\right) .
\end{equation}
Then, the map ${\Lambda\left(\xi, \mathbf{u}\right)}$ is an equivariant lift for the system in \equref{bas} with respect to the defined symmetry group.
\end{theorem}

The lift in \equref{lift_bas} associates the system input with the Lie algebra of the symmetry group allowing the construction of a lifted system on the symmetry group\cit{Mahony2020EquivariantDesign}. let $X \in \grpG$ be the state of the lifted system, and let $\xi_0 \in \calM$ be the selected origin, then the lifted system is written as

\begin{equation}
    \dot{X} = \td L_{X}\Lambda\left(\phi_{\xi_{0}}\left(X\right), \Vector{}{u}{}\right) = X\Lambda\left(\phi_{\xi_{0}}\left(X\right), \Vector{}{u}{}\right).
\end{equation}

%% file: sections/eqf.tex
\section{Equivariant Filter Design}\label{sec:eqf}

\subsection{Equivariant Filter Matrices}

Let ${\Lambda}$ be the equivariant lift defined in \equref{lift_bas}, $\xi_{0}$ be the chosen state origin and ${\hat{X} \in \grpG}$ be the \acl{eqf} state, with initial condition ${\hat{X}\left(0\right) = I}$, the identity element of the symmetry group $\grpG$. Therefore the \acl{eqf} state evolves according to
\begin{equation*}
    \dot{\hat{X}} = \td L_{\hat{X}} \Lambda(\phi_{\xi_0}(\hat{X}), \mathbf{u}) + \td R_{\hat{X}} \Delta = \hat{X}\Lambda(\phi_{\xi_0}(\hat{X}), \mathbf{u}) + \Delta \hat{X},
\end{equation*}
where $\Delta$ is the innovation term of the \acl{eqf}, that depends on the choice of local coordinates\cit{Mahony2020EquivariantDesign}.
The state error in the homogeneous space is defined to be ${e = \phi_{\hat{X}^{-1}}\left(\xi\right) \in \calM}$, and its counterpart in the symmetry group is defined to be ${E = X\hat{X}^{-1} \in \grpG}$. Therefore, to compute the linearized error dynamics we first need to choose an origin $\xi_0$. In what follows, the origin is chosen to be the identity of the homogeneous space, ${\xi_0 = \text{id}}$.
Then, we need to select local coordinates on the homogeneous space, to do so, we need to define a chart and a chart transition map ${\vartheta \AtoB{\mathcal{U}_{\xi_0} \subset \calM}{\R^{6}}}$. The choice of local coordinates is free, however, a natural choice is represented by exponential coordinates, therefore we define
\begin{equation}
\label{eq:local_coords_bas}
    \varepsilon = \vartheta\left(e\right) = \vartheta\left(e_{R},\,e_{b},\, e_{C}\right) = \left(\log\left(e_{R}\right)^{\vee},\, e_{b},\, \log\left(e_{C}\right)^{\vee}\right) \in \R^{9} ,
\end{equation}
with ${\vartheta\left(\xi_0\right) = \mathbf{0} \in \R^{9}}$. The derivation of the linearized error dynamics, following\cit{VanGoor2020EquivariantSpaces}, leads to
\begin{align}
    \dot{\varepsilon} &\approx \mathbf{A}_{t}^{0}\varepsilon - \Fr{e}{\xi_0}\vartheta\left(e\right)\Fr{E}{I}\phi_{\xi_0}\left(E\right)\left[\Delta\right] \label{eq:linearisation_epsilon} ,\\
    \mathbf{A}_{t}^{0} &= \Fr{e}{\xi_0}\vartheta\left(e\right)\Fr{E}{I}\phi_{\xi_0}\left(E\right)\Fr{e}{\xi_0}\Lambda\left(e, \Vector{}{u}{0}\right)\Fr{\varepsilon}{\mathbf{0}}\vartheta^{-1}\left(\varepsilon\right) ,\label{eq:A}
\end{align}
where ${\Vector{}{u}{0} \coloneqq \psi\left(\hat{X}^{-1}, \Vector{}{u}{}\right)  = \left(\Vector{}{\omega}{0},\,\mathbf{0},\,\mathbf{0},\,\mathbf{0}\right)}$ is the origin input. 

The output map in \equref{confout_bas} is already defined on a linear vector space, thus it does not require any choice of local coordinates. Then, following\cit{VanGoor2020EquivariantSpaces}, linearizing the output as a function of the coordinates $\varepsilon$ about $\varepsilon = \mathbf{0}$ yields
\begin{align}
    \Vector{}{y}{} &\approx \mathbf{C}^{0}\bm{\varepsilon} , \label{eq:linearisation_delta} \\
    \mathbf{C}^{0} &= \Fr{e}{\xi_0}h\left(e\right)\Fr{\bm{\varepsilon}}{\mathbf{0}}\varepsilon^{-1}\left(\bm{\varepsilon}\right) .\label{eq:C}
\end{align}

Solving the equations \eqref{eq:A} and \eqref{eq:C} for the linearized error state matrix ${\mathbf{A}_{t}^{0}}$ and the linearized output matrix ${\mathbf{C}^{0}}$ leads to
\begin{subequations}
    \begin{minipage}{0.495\linewidth}
        \begin{equation}
            \mathbf{A}_{t}^{0} = \begin{bmatrix}
                \mathbf{0} & -\mathbf{I} & \mathbf{0}\\
                \mathbf{0} & \Vector{}{\bm{\omega}}{0}^{\wedge} & \mathbf{0}\\
                \mathbf{0} & \mathbf{0} & \Vector{}{\bm{\omega}}{0}^{\wedge}
            \end{bmatrix} , \label{eq:At0_bas}
        \end{equation}
    \end{minipage}
    \begin{minipage}{0.495\linewidth}
        \begin{equation}
            \mathbf{C}^{0} = \begin{bmatrix}
                \Vector{}{d}{1}^{\wedge} & \mathbf{0} & \Vector{}{d}{1}^{\wedge}\\
                \Vector{}{d}{2}^{\wedge} & \mathbf{0} & \mathbf{0}
            \end{bmatrix}. \label{eq:C0_bas}
        \end{equation}
    \end{minipage}
    \vspace{7.5pt}
\end{subequations}

Note that the major difference in the \ac{eqf} filter matrices for the case of an already calibrated sensor compared to an estimated online extrinsic calibration is in the $\mathbf{C}^{0}$ matrix. In particular, the known direction $\Vector{}{d}{}$ appears only in the first block column of the $\mathbf{C}^{0}$ matrix for the case of already calibrated sensors, whereas it appears on both the first block column and on the block column that corresponds to the extrinsic calibration of that specific sensor, in the case of an estimated online extrinsic calibration. 

\subsection{EqF Formulation and Practical Implementation}
\label{sec:eqf_implementation}

To summarize, Let $\hat{X} \in \grpG$ be the \acl{eqf} state with initial condition ${\hat{X}\left(0\right) = I}$. Let ${\bm{\Sigma} \in \PD\left(9\right) \subset \R^{9 \times 9}}$ be the Riccati (covariance) matrix of the error in local coordinates, with initial condition ${\bm{\Sigma}\left(0\right) = \bm{\Sigma}_0}$. Let $\mathbf{A}_{t}^{0}$ and $\mathbf{C}^{0}$ be the matrices defined respectively in \equref{At0_bas} and \equref{C0_bas}. Let ${\bm{M}_c \in \PD\left(9\right) \subset \R^{9 \times 9}}$ be a continuous-time state gain matrix (input covariance).
The continuous-time propagation equations of the \acl{eqf} are given by 
\begin{align}
    &\dot{\hat{X}} = \hat{X}\Lambda(\phi_{\xi_0}(\hat{X}), \mathbf{u}) \label{eq:eqf_ode_bas},\\
    & \dot{\bm{\Sigma}} = \mathbf{A}_{t}^{0}\bm{\Sigma} + \bm{\Sigma}{\mathbf{A}_{t}^{0}}^T + \bm{M}_c,\label{eq:riccatiode}
\end{align}

A discrete-time implementation of the filter's propagation phase requires integration of \equref{eqf_ode_bas}-(\ref{eq:riccatiode}) in the time-step $\Delta T$ in between gyro measurements\cit{maybeck1982stochastic},\cit{Trawny2005IndirectAlgebra}. The \ac{eqf} state-transition matrix ${\mathbf{\Phi}\left(t+\Delta t, t\right) = \mathbf{\Phi}}$ can be written as
\begin{equation}\label{eq:stm}
    \mathbf{\Phi} = \begin{bmatrix}
    \eye & \mathbf{\Phi}_{12} & \mathbf{0}\\
    \mathbf{0} & \mathbf{\Phi}_{22} & \mathbf{0}\\
    \mathbf{0} & \mathbf{0} & \mathbf{\Phi}_{22}
    \end{bmatrix}
\end{equation}
with
\begin{align*}
    \mathbf{\Phi}_{12} &\simeq -\Delta T\left(\eye + \frac{\Delta T}{2}\Vector{}{\bm{\omega}}{0}^{\wedge} + \frac{\Delta T^2}{6}\Vector{}{\bm{\omega}}{0}^{\wedge}\Vector{}{\bm{\omega}}{0}^{\wedge}\right),\\
    \mathbf{\Phi}_{22} &\simeq \eye + \Delta T\Vector{}{\bm{\omega}}{0}^{\wedge} + \frac{\Delta T^2}{2}\Vector{}{\bm{\omega}}{0}^{\wedge}\Vector{}{\bm{\omega}}{0}^{\wedge}.
\end{align*}

Therefore, the \ac{eqf}'s propagation phase is implemented in discrete-time according to the following equations
\begin{align}
    &\hat{X}_{k+1}^{-} = \hat{X}_{k}^{+} + \int_{0}^{\Delta T} \hat{X}\left(t\right)\Lambda(\phi_{\xi_0}(\hat{X}\left(t\right)), \mathbf{u}\left(t\right)) \; dt, \label{eq:meanprop}\\
    &\bm{\Sigma}_{k+1}^{-} = \mathbf{\Phi}\bm{\Sigma}_{k}^{+}\mathbf{\Phi}^T + \bm{M}_d, \label{eq:covprop}\\
    &\bm{M}_d = \int_{t}^{t + \Delta T} \mathbf{\Phi}\left(t+\Delta T, \tau\right)\bm{M}_c\mathbf{\Phi}^T\left(t+\Delta T, \tau\right) \; \td \tau, \label{eq:Md_eqf}
\end{align}
where the mean integration can be done numerically (e.g. with Lie group integrator or RK4 schemes). 
and $\bm{M}_d$ can be approximated by ${\bm{M}_d \simeq \bm{M}_c\Delta T}$\cit{Sola2017QuaternionFilter}.

Let ${\bm{N} \in \PD\left(6\right) \subset \R^{6 \times 6}}$ be an output gain matrix (measurement covariance). 
The \acl{eqf} update equations are 
\begin{align}
    &\mathbf{S} = \mathbf{C}^{0}\bm{\Sigma}_{k+1}^{-}{\mathbf{C}^{0}}^T + \bm{N}, \label{eq:dtS}\\
    &\mathbf{K} = \bm{\Sigma}_{k+1}^{-}{\mathbf{C}^{0}}^T\mathbf{S}^{-1}, \label{eq:dtK}\\
    &\Delta = \Fr{E}{I}\phi_{\xi_0}\left(E\right)^{\dagger}\td\varepsilon^{-1}\mathbf{K}\rho_{\hat{X}^{-1}}\left(\Vector{}{y}{}\right), \label{eq:dtinno}\\
    &\hat{X}_{k+1}^{+} = \exp\left(\Delta\right)\hat{X}_{k+1}^{-}, \label{eq:dtupdate}\\
    &\bm{\Sigma}_{k+1}^{+} = \left(\eye - \mathbf{K}\mathbf{C}^0\right)\bm{\Sigma}_{k+1}^{-}\label{eq:dtsigupdate}. 
\end{align}

\begin{remark}
    A major difference with respect to the commonly known Kalman Filter equations, is \equref{dtinno}, where the equivariant residual ${\rho_{\hat{X}^{-1}}\left(\Vector{}{y}{}\right)}$ gets scaled by the Kalman gain ${\mathbf{K}}$ and then mapped back to the Lie algebra of the symmetry group. This innovation term is then used to update the propagated state estimate in \equref{dtupdate}.
\end{remark}
Furthermore, note that the estimate of the original system $\hat{\xi}$ can be retrieved from the lifted system estimate $\hat{X}$ with the following relation ${\hat{\xi} = \phi_{\hat{X}}\left(\xi_0\right)}$.

A practical implementation of the proposed filter needs to handle noise in the input variables and in the output measurements, and hence the gain matrices $\bm{M}_c$ and $\bm{N}$ need to be computed according to the covariance matrices of the input and measurement noise respectively. Let $\mathbf{u}_{m} = \mathbf{u} + \mathbf{u_n}$ model the measured input, where $\mathbf{u_n}$ is additive white Gaussian noise with covariance ${\bm{\Sigma}_{u}}$. Therefore the equivariant lift presented in \equref{lift_bas} as well as the origin input, change accordingly to $\Lambda\left(\xi, \mathbf{u}\right) + \Lambda_l\left(\xi, \bm{u_n}\right)$, and ${\Vector{}{u}{0} + \Vector{}{u_n}{0} = \psi\left(\hat{X}^{-1}, \Vector{}{u}{}\right) + \psi_l\left(\hat{X}^{-1}, \Vector{}{u_n}{}\right)}$, where ${\Lambda_l\left(\xi, \bm{u_n}\right)}$ and ${\psi_l\left(\hat{X}^{-1}, \Vector{}{u_n}{}\right)}$ are the linear part of the lift ${\Lambda}$ and the action $\psi$ applied to the measurement noise. By considering measurement noise, the derivation of the linearized error dynamics changes with respect to \equref{linearisation_epsilon}, in particular, following the steps in\cit{VanGoor2020EquivariantSpaces} and including the previously defined noise term, leads to
\begin{align*}
    \dot{e} &= \td\phi_{e}\left(\Lambda\left(e, \mathbf{u}_0 - \Vector{}{u_n}{0} \right) - \Lambda\left(\xi_0, \mathbf{u}_0\right) - \Delta\right)\\
    &= \td\phi_{e}\left(\Lambda\left(e, \mathbf{u}_0\right) - \Lambda\left(\xi_0, \mathbf{u}_0\right)\right) - \td\phi_{e}\Delta - \td\phi_{e}\Lambda_l\left(e,\,\Vector{}{u_n}{0}\right)\\
    &= \td\phi_{e}\Tilde{\Lambda}_{\xi_0}\left(e, \mathbf{u}_0\right) - \td\phi_{e}\Delta - \td\phi_{e}\Lambda_l\left(e,\,\Vector{}{u_n}{0}\right) ,
\end{align*}
therefore, linearizing the error dynamics in the local coordinates $\bm{\varepsilon}$, defined in \equref{local_coords_bas}, around ${\bm{\varepsilon} = \mathbf{0}}$ results in\cit{Mahony2021ObserverEquivarianceb}
\begin{align}
    \dot{\bm{\varepsilon}} &\approx \mathbf{A}_{t}^{0}\bm{\varepsilon} + \mathbf{B}_{t}^{0}\Vector{}{u_n}{} - \Fr{e}{\xi_0}\varepsilon\left(e\right)\Fr{E}{I}\phi_{\xi_0}\left(E\right)\left[\Delta\right] \label{eq:linearization_epsilon_with_noise},\\
    \mathbf{B}_{t}^{0} &= \td\varepsilon\td\phi_{e}\td\Lambda_l\left(\xi_0,\,\Vector{}{u_n}{0}\right) \label{eq:B}. 
\end{align}
Similarly, let $\Vector{}{y}{}+\mathbf{n}$ be a real-world attitude measurement, where $\mathbf{n}$ is additive white Gaussian noise with covariance ${\bm{\Sigma}_{y}}$, then it is trivial to note that ${h\left(e\right)}$ can be written 
\begin{equation*}
    \begin{split}
        h\left(e\right) &= \rho_{\hat{X}^{-1}}\left(h\left(\xi\right)\right) = \rho_{\hat{X}^{-1}}\left(\Vector{}{y}{} -\Vector{}{n}{}\right)
        = \rho_{\hat{X}^{-1}}\left(\Vector{}{y}{}\right) - \rho_{\hat{X}^{-1}}\left(\Vector{}{n}{}\right),
    \end{split}
\end{equation*}

The gain matrices $\bm{M}_c$ and $\bm{N}$ are written according to
\begin{equation*}
    \bm{M}_c =  \mathbf{B}_{t}^{0}\bm{\Sigma}_u\mathbf{B}_{t}^{0^T} \qquad\qquad
    \bm{N} =  \mathbf{D}_{t}^{0}\bm{\Sigma}_y\mathbf{D}_{t}^{0^T} ,
\end{equation*}
with
\begin{equation*}
    \mathbf{B}_{t}^{0} = \begin{bmatrix}
        \hat{A} & \mathbf{0} & \mathbf{0}\\
        \mathbf{0} & \hat{A} & \mathbf{0}\\
        \mathbf{0} & \mathbf{0} & \hat{B}
    \end{bmatrix}
    \qquad\qquad
    \mathbf{D}_{t}^{0} = \begin{bmatrix}
        \hat{B} & \mathbf{0}\\
        \mathbf{0} & \hat{A}
    \end{bmatrix} .
\end{equation*}

%% file: sections/comparison.tex
\section{Simulated Experiments and Comparison}\label{sec:comp}

In this section we compare the proposed \acl{eqf} design based the proposed symmetry with the Imperfect-\ac{iekf}\cit{barrau:tel-01247723},\cite{doi:10.1177/0278364919894385} (referred as \ac{iekf} from now on, in text, plots and tables for improved readability). 
In the interest of fairness, we limit our comparison to filtering approaches, and in particular to the Imperfect-\ac{iekf} only since it is the state-of-the-art filter solution for navigation problem and is proven to outperform other filter\cit{doi:10.1177/0278364919894385}.
In this section the comparison is done on a simulated experiment of an \ac{uav} equipped with a gyroscope and receiving body-frame direction measurements from a magnetometer ($\Vector{}{y}{1}$), and spatial direction measurements of from two \ac{gnss} receivers ($\Vector{}{y}{2}$). Measurements $\Vector{}{y}{1}$ and $\Vector{}{y}{2}$ are defined accordingly to the measurement model in \equref{confout_bas} given the directions $\Vector{}{d}{1}$ and $\Vector{}{d}{2}$. Ground-truth data is generated through the Gazebo/RotorS framework\cit{furrer_rotors_2016} that simulates a \ac{uav}’s flight and sensor behavior by realistically modeling the \ac{uav}’s dynamics and sensor measurements. Using this simulation setup has the advantage of having access to ground-truth data for all states of interest for the latter evaluation and high repeatability of the conducted experiments.

\subsection{Background}
It is of particular interest to understand how we construct a spatial direction measurement of a body-frame reference direction, from the position measurements of two \ac{gnss} receivers. The platform is equipped with two \ac{gnss} receivers, $\Vector{}{r}{1}$ and $\Vector{}{r}{2}$, placed with sufficient baseline between each other (eg. as in \figref{twins_plots}). At the intersection of the baseline between the \ac{gnss} receivers and one of the three planes spanned by the body frame axes, we place a "virtual" \ac{gnss} frame \frameofref{g} with the y-axis aligned along the baseline. 
In our setup, the "virtual" frame \frameofref{g} overlaps with the body frame \frameofref{I} (i.e., $\Rot{I}{g}=\eye$), and the baseline between the \ac{gnss} receivers is set to $1$\si[per-mode = symbol]{\meter}. 
We model the body-frame direction measurement as ${\Vector{}{y}{2} = \left[0\,1\,0\right]^T}$, resulting from the measurement model in \equref{confout_bas} with a time varying spatial reference direction $\Vector{}{d}{2}$, given by the rotation of the y-axis of the \frameofref{g} frame into the global inertial frame \frameofref{G}. The spatial reference direction $\Vector{}{d}{2}$ can be easily constructed from the raw \ac{gnss} measurements as follows
\begin{equation}\label{eq:gps_dir}
    \Vector{}{d}{2} = \frac{\left(\Vector{G}{p}{r_1}-\Vector{G}{p}{r_2}\right)}{\norm{\left(\Vector{G}{p}{r_1}-\Vector{G}{p}{r_2}\right)}}.
\end{equation}

A reference algorithm for the proposed \ac{eqf} is shown in \algref{eqf}. Note that the EqF state ${\hat{X}}$ is implemented as two matrices. ${\left(\hat{A},\, \hat{a}\right) = \begin{bsmallmatrix}\hat{A} & \hat{a}^{\vee}\\\mathbf{0} & 1\end{bsmallmatrix}}$, and $\hat{B}$ as a $3 \times 3$ rotation matrix.
${\exp}$ refers to the matrix exponential. The notation ${\Vector{}{r}{i:k}}$ refers to the sub-vector of ${\Vector{}{r}{}}$ with starting index $i$, and ending index $k$. A detailed version is included in the report\cit{supp_seafile}.

\newcommand\mycommfont[1]{\scriptsize\ttfamily{#1}}
\SetCommentSty{mycommfont}
\SetKwProg{Loop}{loop}{}{end}
\SetKwProg{Propagation}{propagation}{}{end}
\SetKwProg{Update}{update}{}{end}
\SetInd{0.1em}{0.25em}
\begin{algorithm}[!htp]
\caption{Proposed EqF (main loop)}\label{alg:eqf}
\KwIn{Gyro measurement: ${\bm{\omega}}$}
\Propagation{}{
$\Vector{}{\omega}{0} \coloneqq \hat{A}\Vector{}{\omega}{} + \hat{a}^{\vee}$\;
$\begin{bsmallmatrix}
    \hat{A} & \hat{a}^{\vee}\\
    \mathbf{0} & 1
\end{bsmallmatrix} \gets 
\begin{bsmallmatrix}
    \hat{A} & \hat{a}^{\vee}\\
    \mathbf{0} & 1
\end{bsmallmatrix}\exp\left(\begin{bsmallmatrix}
    \left(\hat{A}^T\Vector{}{\omega}{0}\right)^{\wedge} & \Vector{}{\omega}{}^{\wedge}\hat{A}^T\hat{a}^{\vee}\\
    \mathbf{0} & 0
\end{bsmallmatrix}\Delta t\right)$\tcp*{\ref{eq:meanprop}}
$\hat{B} \gets \hat{B}\exp\left(\hat{B}^T\hat{A}\left(\hat{A}^T\Vector{}{\omega}{0}\right)^{\wedge}\hat{A}^T\hat{B}\Delta t\right)$\tcp*{\ref{eq:meanprop}}
$\bm{\Sigma} \gets \mathbf{\Phi}\bm{\Sigma}\mathbf{\Phi}^T + \mathbf{B}_{t}^{0}\bm{\Sigma}_{u}{\mathbf{B}_{t}^{0}}^T\Delta t$\tcp*{\ref{eq:covprop} with ${\bm{\Phi}}$ in \ref{eq:stm}}
}
\KwIn{Magnetometer direction measurement: ${\Vector{}{y}{1}}$}
\KwIn{GNSS direction: ${\Vector{}{d}{2}}$ (\ref{eq:gps_dir})}
\Update{}{
$\mathbf{K} = \bm{\Sigma}{\mathbf{C}^{0}}^T\left(\mathbf{B}^{0}\bm{\Sigma}{\mathbf{C}^{0}}^T + \mathbf{D}_{t}^{0}\bm{\Sigma}_{y}{\mathbf{D}_{t}^{0}}^T\right)^{-1}$\tcp*{\ref{eq:dtK}}
$\Vector{}{r}{} = \mathbf{K}\begin{bsmallmatrix}
    \hat{B}\Vector{}{y}{1}\\
    \hat{A}\Vector{}{y}{2}
\end{bsmallmatrix}$\tcp*{Part of \ref{eq:dtinno} with $\Vector{}{y}{2} =  \left[0\,1\,0\right]^T$}
$\begin{bsmallmatrix}
    \hat{A} & \hat{a}^{\vee}\\
    \mathbf{0} & 1
\end{bsmallmatrix} \gets 
\exp\left(\begin{bsmallmatrix}
    \Vector{}{r}{1:3}^{\wedge} & -\Vector{}{r}{4:6}\\
    \mathbf{0} & 0
\end{bsmallmatrix}\right)
\begin{bsmallmatrix}
    \hat{A} & \hat{a}^{\vee}\\
    \mathbf{0} & 1
\end{bsmallmatrix}$\tcp*{\ref{eq:dtupdate}}
$\hat{B} \gets \exp\left(\left(\Vector{}{r}{7:9}+\Vector{}{r}{1:3}\right)^{\wedge}\right)\hat{B}$\tcp*{\ref{eq:dtupdate}}
$\bm{\Sigma} \gets \left(\eye - \mathbf{K}\mathbf{C}^0\right)\bm{\Sigma}$\tcp*{\ref{eq:dtsigupdate}}
}
\end{algorithm}
\vspace{-1em}

\subsection{Method}
In the simulation setup, the performance of the \ac{eqf} and \ac{iekf} were evaluated on a 100-runs Monte-Carlo simulation consisting of different \SI{70}{\second} long Lissajous trajectories with different levels of excitation. In each run the two filters were randomly initialized with wrong attitude (${10}$\si[per-mode = symbol]{\degree} error std per axis), identity calibration, and zero bias. The initial covariance is set large enough to cover the initialization error.
In order to replicate a realistic scenario, simulated gyroscope measurements provided to the filters at \SI{200}{\hertz} included Gaussian noise and a non zero, time-varying bias (modeled as a random walk process). The continuous-time standard deviation of the measurement noise are ${\sigma_{\Vector{}{w}{}} = 8.73\cdot10^{-4}}$\si[per-mode = symbol]{\radian\per\sqrt\second}, and ${\sigma_{\Vector{}{b}{w}} = 1.75\cdot10^{-5}}$\si[per-mode = symbol]{\radian\per\second\sqrt\second} for simulated gyroscope measurements and bias respectively. Zero mean white Gaussian noise was added to the two direction measurements $\Vector{}{y}{1}$, and $\Vector{}{y}{2}$ and they were provided to the filters at \SI{100}{\hertz}, and \SI{20}{\hertz} respectively. The discrete-time standard deviation\footnote{The  discrete-time standard deviation ${\sigma_{\Vector{}{y}{2}}}$ is derived simulating two \ac{rtk} \ac{gnss} receivers with high position accuracy (e.g. with a standard deviation of ${\sigma_{\Vector{G}{p}{r}} = 0.1}$\si[per-mode = symbol]{\meter} in each position axis).} of the (unit-less) direction measurement noise are ${\sigma_{\Vector{}{y}{1}} = 0.2}$, and ${\sigma_{\Vector{}{y}{2}} = 0.1}$.
For fair comparison, the gain matrices $\bm{M}_c$ and $\bm{N}$ on both filters were first set to reflect the measurement noise covariance and are only then adapted as discussed in Section \ref{sec:eqf_implementation}.

\begin{figure}[!t]
\centering
\includegraphics[width=0.95\linewidth]{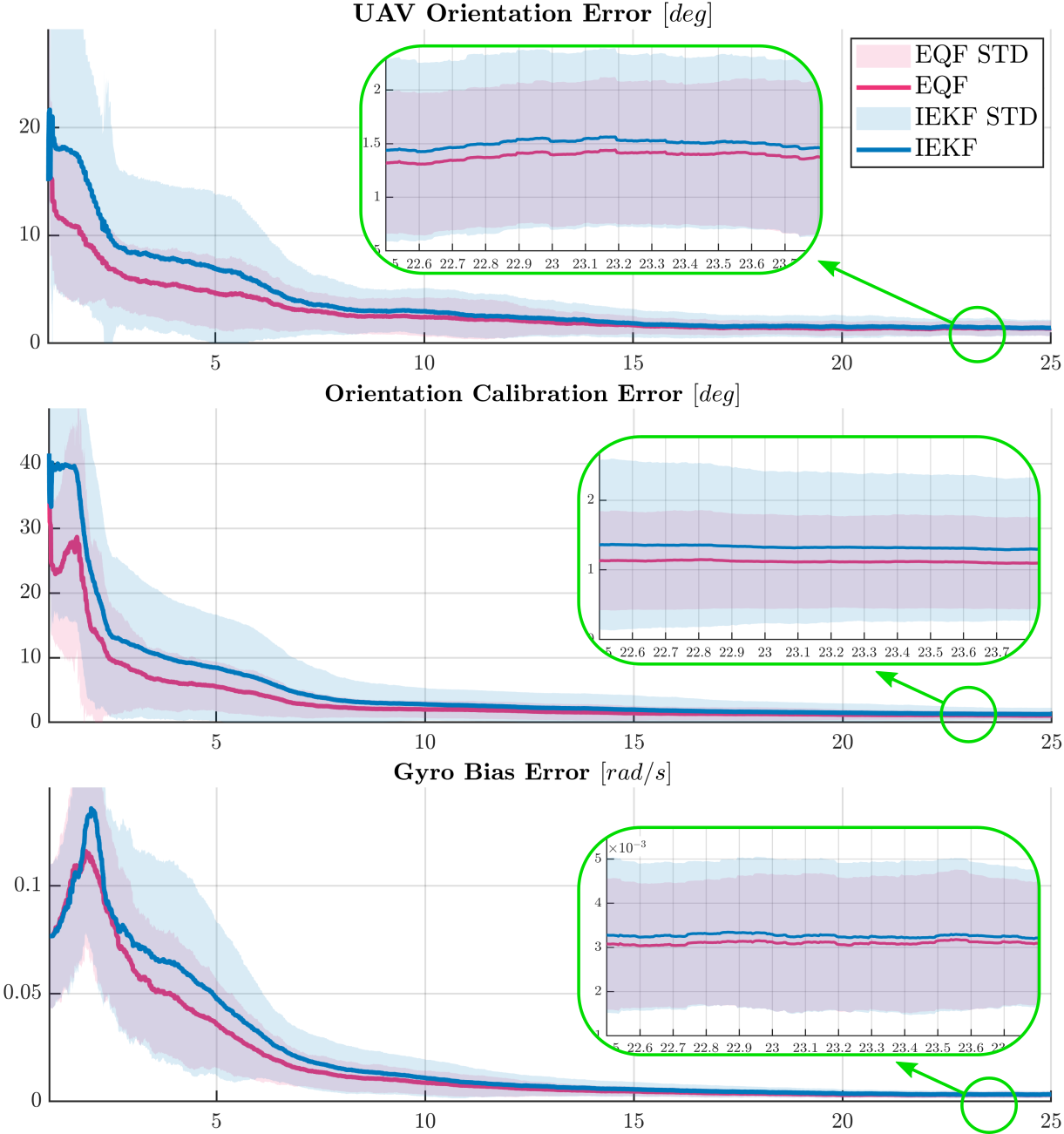}
\vspace{-5pt}
\caption{Averaged norm of the \ac{uav} attitude, magnetometer calibration, and gyroscope bias error over the 100 runs. Note that the plots only depict the first \SI{25}{\second} of the \SI{70}{\second} trajectories showing the improved transient of the proposed \ac{eqf}.
The analysis of the full trajectories is reflected in the numbers in \tabref{rmse}.}
\label{fig:error_plots}
\vspace{-5pt}
\end{figure}
\begin{table}[!t]
\renewcommand{\arraystretch}{1.15}
\setlength\tabcolsep{4.5pt}
\centering
\scriptsize
\caption{MC simulation transient and asymptotic averaged RMSE}
\vspace{-5pt}
\begin{tabular}{c c c c c c}
\specialrule{.1em}{.1em}{.1em} 
\textsc{Rmse} & Attitude [\si[per-mode = symbol]{\degree}]& bias [\si[per-mode = symbol]{\radian\per\second}] & Calibration [\si[per-mode = symbol]{\degree}] \\ \hline
\ac{eqf} $(T)$ & $\mathbf{3.5331}$ & $\mathbf{0.0280}$ & $\mathbf{5.7892}$ \\
\acs{iekf} $(T)$ & $4.9497$ & $0.0323$ & $8.0480$\\
\specialrule{.1em}{.1em}{.1em} 
\ac{eqf} $(A)$ & $\mathbf{1.3870}$ & $\mathbf{0.0035}$ & $\mathbf{0.6989}$ \\
\acs{iekf} $(A)$ & $1.3995$ & $\mathbf{0.0035}$ & $0.7798$\\
\specialrule{.1em}{.1em}{.1em} 
\end{tabular}
\label{tab:rmse}
\vspace{-10pt}
\end{table}

\subsection{Discussion}
\tabref{rmse} reports the averaged \acs{rmse} of the filter states over the 100 runs. In the table, $T$ denotes that the \acs{rmse} is computed for the transient phase considered the first \SI{35}{\second} whereas $A$ denotes that the \acs{rmse} is computed for the asymptotic phase considered the last \SI{35}{\second} of each run. \figref{error_plots} shows the evolution of the averaged filter states error and its sample standard deviation over the first \SI{25}{\second} of the \SI{70}{\second} trajectories for a better overview over the transient phase. In general, it is interesting to note the improved tracking and transient performance of the proposed \ac{eqf}, and the ability of the filter to converge quickly despite the heavily wrong initializations (the mean error norms for both filter start at roughly \SI{20}{\degree} and \SI{35}{\degree} for the orientation and calibration states respectively).
The \ac{eqf} derived from our proposed symmetry outperforms the state-of-the-art \ac{iekf} in both transient response and asymptotic behaviour although the simulated Gaussian noise, high rates, and uninterrupted measurements are bread and butter for the bias states "tacked on" in the \ac{iekf} formulation. Non-ideal conditions as shown below in the real-world tests make the differences between the approaches even more apparent.

\subsection{Runtime}
State independencies and several (partly-)constant matrices as discussed in Section \ref{sec:eqf} may give the impression that runtime is a non-issue for our proposed \ac{eqf}. However, the continuous time formulation triggers questions on the practical implementation approach. 
\tabref{runtime} reports a comparison of the runtime for four different approaches.
(i) The proposed analytically derived closed form $\bm{\Phi}$ as in \equref{stm}. ii) An equivalent ${\bm{\Phi} = \exp\left(\mathbf{A}_{t}^{0} \Delta t\right)}$ solved numerically at each iteration ($\mathbf{A}_{t}^{0}$ as in \equref{At0_bas}). iii) A first order approximation ${\bm{\Phi} = \eye + \mathbf{A}_{t}^{0} \Delta t}$, where the Riccati equation \equref{riccatiode} is integrated with first order truncated Euler method. iv) Matlab's ode45 solution of the continuous-time equations \eqref{eq:eqf_ode_bas}-\eqref{eq:riccatiode}.
All four implementations were run in Matlab on the same system and on the same data multiple times, and the final execution time for each of the aforementioned cases were averaged. While (i) and (ii) are equivalent in terms of accuracy, results in \tabref{runtime}, show that the proposed \ac{eqf} implementation is faster. When compared to (iii), the proposed closed form solution avoids approximations on the covariance propagation, while preserving the runtime, showing the importance of a proper discrete time filter derivation for practical implementations.
\begin{table}[!t]
\renewcommand{\arraystretch}{1.15}
\setlength\tabcolsep{4.5pt}
\centering
\scriptsize
\caption{Relative runtime comparison in percentage}
\vspace{-5pt}
\begin{tabular}{c c c c}
\specialrule{.1em}{.1em}{.1em} 
(i) \textsc{Proposed} & (ii) $\bm{\Phi} = \exp\left(\mathbf{A}_{t}^{0} \Delta t\right)$ & (iii) $\bm{\Phi} = \eye + \mathbf{A}_{t}^{0} \Delta t$ & (iv) \textsc{ode45} \\ \hline
$100\%$ & $125\%$ & $102\%$ & $>1000\%$ \\
\specialrule{.1em}{.1em}{.1em} 
\end{tabular}
\label{tab:runtime}
\vspace{-15pt}
\end{table}

%% file: sections/experiment.tex
\section{Indoor Real-World Experiments}\label{sec:exp_indoor}

In this section, we compare against accurate ground truth in a real scenario of multi-rate sensors, unsynchronized measurements, and measurement dropouts, showing that the proposed \ac{eqf} outperforms the state-of-the-art \ac{iekf}, and that it is suitable for real-world (non-ideal) sensor data.

\subsection{Method}
The indoor dataset was recorded with an AscTec Hummingbird \ac{uav} flying an aggressive trajectory in a motion capture equipped room for \SI{140}{\second}. Gyroscope measurements, as well as full 6-DOF pose measurements of the platform were available at \SI{330}{\hertz}. The continuous-time standard deviation of the gyroscope measurement and bias noise are set to  ${\sigma_{\Vector{}{w}{}} = 0.013}$\si[per-mode = symbol]{\radian\per\sqrt\second}, and ${\sigma_{\Vector{}{b}{w}} = 0.0013}$\si[per-mode = symbol]{\radian\per\second\sqrt\second} respectively.
The pose measurements from the motion capture system were used to re-create the previously discussed scenario, and therefore to manufacture measurements $\Vector{}{y}{1}$ and $\Vector{}{y}{2}$ of directions $\Vector{}{d}{1}$ and $\Vector{}{d}{2}$. To replicate the non-idealities of real-world measurements, $\Vector{}{y}{1}$, and $\Vector{}{y}{2}$ were generated at \SI{100}{\hertz} and \SI{25}{\hertz} respectively. A dropout rate of $\sim10\%$ was actively induced for the magnetometer measurements. Moreover the measurements were then perturbed with zero mean white Gaussian noise with (unit-less) discrete-time standard deviations ${\sigma_{\Vector{}{y}{1}} = 0.1}$, and ${\sigma_{\Vector{}{y}{2}} = 0.01}$. 
The two filters were both initialized with an wrong attitude that correspond to the Euler angles (ypr) ${\hatVector{G}{\theta}{}\simeq\left[70\;-40\;30\right]^T}$ in degrees, and calibration corresponding to the Euler angles (ypr) ${\hatVector{I}{\phi}{}=\left[-90\;-60\;130\right]^T}$ in degrees, and zero bias. The ground-truth initial attitude corresponds to the Euler angles (ypr) ${\Vector{G}{\theta}{}\simeq\left[90\;0\;0\right]^T}$ in degrees, and the ground-truth extrinsic calibration corresponds to the Euler angles (ypr) ${\Vector{I}{\phi}{}=\left[30\;5\;25\right]^T}$ in degrees. The initial covariance is set large enough to cover the initialization error. For fair comparison, the two filters shared the same gain tuning prior to filter-specific adaptation (cf. Sections \ref{sec:eqf_implementation}).

\subsection{Discussion}
\begin{figure}[!t]
\centering
\includegraphics[width=0.95\linewidth]{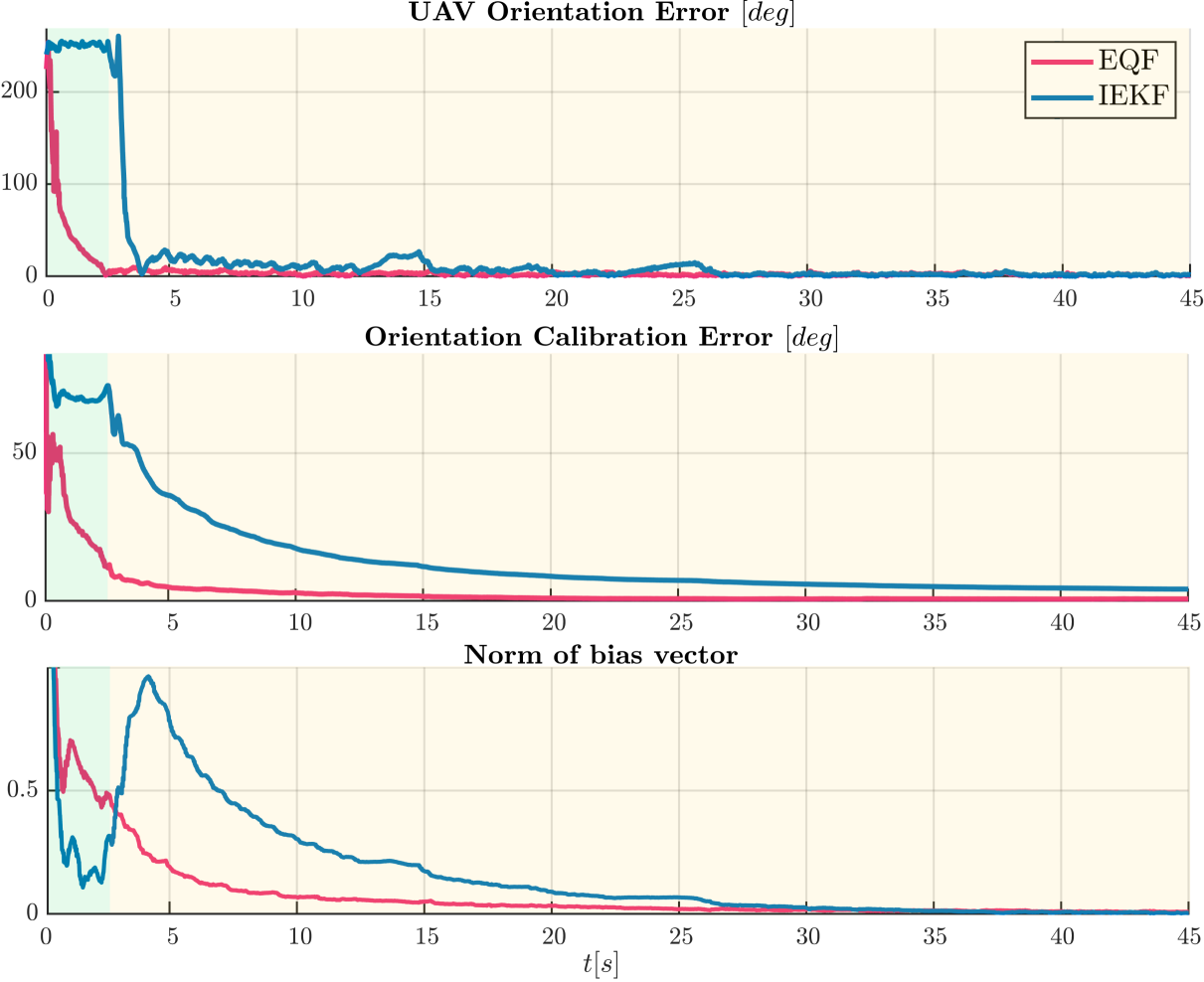}
\vspace{-10pt}
\caption{Norm of the \ac{uav} attitude, magnetometer calibration error, and norm of the estimated bias vector for the indoor experiments. 
Note that, in contrast to the \ac{eqf}, only after \SI{2.5}{\second} is the \ac{uav} providing sufficient angular excitation (i.e. mission phase, yellow shaded, after vertical take-off, green shaded) for the \ac{iekf} to extract information.
}
\label{fig:hummy_error_plots}
\vspace{-15pt}
\end{figure}

In this indoor experiment, we have specifically chosen to initialize the filters with completely wrong attitude and calibration states to trigger the worst possible scenario in order to highlight the estimator's transient behaviors. 
\figref{hummy_error_plots} shows the transients of the attitude error, extrinsic calibration error, and the bias norm for the first \SI{45}{\second} of the trajectory. In this experiment, the proposed \ac{eqf} clearly outperforms the \ac{iekf}, in particular, for the \ac{eqf} the attitude error norm stabilizes below \SI{10}{\degree} in $\sim$\SI{3}{\second}, and below \SI{5}{\degree} in $\sim$\SI{10}{\second}, while the \ac{iekf} requires respectively $\sim$\SI{15}{\second}, and $\sim$\SI{30}{\second}. In terms of calibration states error, the proposed \ac{eqf} shows a transient almost an order of magnitude faster, converging below \SI{5}{\degree} in just $\sim$\SI{5}{\second} compared to the $\sim$\SI{50}{\second} of the\ac{iekf}. Although no ground-truth information is available for the gyroscope bias we compared the norm of the bias vector for the two filters. The rationale behind such comparison is that the norm of the bias vector is supposed to quickly decrease almost to zero since the AscTec platforms perform a gyro calibration upon system start. Results in \figref{hummy_error_plots} show a faster decrement of that norm for the proposed \ac{eqf} compared to the \ac{iekf} and therefore a faster convergence of the bias state is assumed. In this flight, the \ac{uav} was taking-off vertically for the first 2.5 seconds lowering the amount of information in the input (low angular velocities). While the \ac{iekf} struggles to extract information for the state estimation the \ac{eqf} is able to do so even in this low signal-to-noise ratio situation.

\section{Outdoor Real World Experiments}\label{sec:exp_outdoor}

Although no ground truth information is available for this outdoor experiment, we show an application of the filters in a realistic scenario with realistic sensor data from a magnetometer and two \ac{rtk} \ac{gnss} receiver (as opposed to constructed measurements from a tracking system as done for the indoor tests).

\subsection{Method}
The outdoor dataset was recorded by flying a TWINS Science quadrotor for \SI{130}{\second}, equipped with a PixHawk4 FCU with \acs{imu}, magnetometer and two Ublox \ac{rtk} \ac{gnss} receivers as shown in \figref{twins_plots}. Gyroscope measurements were available at \SI{200}{\hertz}, magnetometer (calibrated for soft, and hard-iron effects), and \ac{gnss} measurements were available at \SI{90}{\hertz} and \SI{8}{\hertz} respectively. The continuous-time standard deviation of the gyroscope measurement noise are ${\sigma_{\Vector{}{w}{}} = 1.75\cdot10^{-4}}$\si[per-mode = symbol]{\radian\per\sqrt\second}, and ${\sigma_{\Vector{}{b}{w}} = 8.73\cdot10^{-6}}$\si[per-mode = symbol]{\radian\per\second\sqrt\second} (obtained with Allan variance technique). Discrete-time standard deviations of the (unit-less) direction measurements were considered to be ${\sigma_{\Vector{}{y}{1}} = 0.1}$, and ${\sigma_{\Vector{}{y}{2}} = 0.01}$.
Again, the two filters' gain matrices, prior to adaptation, were set according to the measurement noise. 
In this experiment, we simulate a scenario of mid-air filter re-initialization, from a completely wrong initial estimate (in absence of prior information). Therefore, both filters were initialized with a wrong consecutive rotation (ypr) of $\sim$\SI{30}{\degree} along each single axis for both attitude and calibration states, and with initial bias set to zero. 
Ground-truth information is not available for this experiment, however, we know that the magnetometer extrinsic calibration is almost identity and, as baseline for the attitude, we refer to the estimates of an online available state-of-the-art modular multi-sensor fusion framework, MaRS\cit{Brommer2021MaRS:Framework}, obtained by feeding \ac{imu}, \ac{gnss} positions and velocities as sensor measurements.

\subsection{Discussion}
This experiment shows an application of the proposed \ac{eqf} to a scenario where real sensor readings from a magnetometer and two \ac{rtk} \ac{gnss} receivers were used.
Interesting aspects of this experiment are the poor quality of the magnetometer readings as well as the reduced amount of excitation the platform was subjected to.
Only three noticeable rotations were performed during the flight: a small combined rotation at about second~$5$, and two yaw-only rotations of \SI{180}{\degree} at about seconds~$25$, and~$45$. 
This low angular excitation is a challenging situation for the attitude estimation problem including calibration and bias states.
\figref{twins_plots} shows the filters' attitude error, as well as the norm of the orientation calibration angular vector and the norm of the bias vector for the first \SI{50}{\second} of the trajectory. The attitude error is computed against the estimate of MaRS, previously initialized and converged (for $t\leq0$ with reference to \figref{twins_plots}). The rationale behind comparing the norm of the aforementioned vectors is that both the extrinsic calibration angular vector and the gyroscope bias vector are almost zero and therefore we expect the norm of the estimate of these vectors to decrease. This provides a good measure of transient performance of the filter although the bias and calibration graphs are not useful for indicating asymptotic performance. 
Both filters' performance showed in \figref{twins_plots} are quite remarkable indeed. 
The lack of ground truth and the low excitation make a quantitative analysis of these results challenging. That said, \figref{twins_plots} clearly shows qualitatively the ability of the proposed \ac{eqf} in real-world applications to quickly converge when initialized with a wrong estimate, in quasi-stationary regime, providing a faster and more reliable estimate than the state-of-the-art.

%% file: sections/conclusion.tex
\section{Conclusion}\label{sec:conc}

In this paper we proposed a novel equivariant symmetry that allows an integrated state formulation to incorporate attitude, extrinsic calibration for all added direction sensors, and gyroscope bias states in a single unified geometric structure. Based on the new symmetry, a novel \acl{eqf} design for attitude estimation of an autonomous systems with $N$ direction measurements was presented.  
Our formulation can include an arbitrary number $N$ of generic direction measurements, being either body-frame fixed or spatial directions measurements covering the majority of real-world direction-sensor modalities.
In addition, we proposed a discretization approach such that the continuous time system formulation can be implemented for real-world usage. We showed that our approach is much faster than elaborated discretized integration schemes and even slightly faster than simple Euler integration while avoiding any approximation on the covariance propagation.

Statistically relevant results from simulated as well as real-world experiments (indoor and outdoors) show that the proposed \ac{eqf} outperforms the state-of-the-art Imperfect-\ac{iekf} in both transient and asymptotic tracking performance. The presented results are of particular relevance when considering the practical applicability of the presented \ac{eqf} to real-world scenarios, and the importance of having estimators that are able to converge quickly, without any prior knowledge, and in presence of high initial errors.
Furthermore, multi-rate sensors, unsynchronized measurements, outliers, measurement dropouts present in real-world data are often the cause of poor filter performance. The proposed discrete-time \ac{eqf} implementation based on our proposed symmetry, exhibited robust performance against the non-idealities of such real-world sensor data. The noticeably lower performance for the Imperfect-\ac{iekf} in these real-world scenarios is assumed to stem from the "tacked on" bias states not properly considered in the overall symmetry and losing the group-affine property. This paper shows that exploiting the underlying symmetry of equivariant systems is of paramount importance to design better estimators.

%% file: sections_supp/notation.tex
\section*{Notation}

The following table summarize the notation used throughout the paper

\begin{table}[!h]
    \footnotesize
    \renewcommand{\arraystretch}{1.5}
    \centering
    \caption{Notation description table}
    \vspace{0.25em}
    \begin{tabular}{c c c}
        \specialrule{.1em}{.1em}{.1em} 
        Description & Descriptive notation & Lean notation \\ \hline
        Rigid body orientation & $\Rot{G}{I}$ & $\Rot{}{}$ \\
        Direction sensor's calibration & $\Rot{I}{S}$ & $\mathbf{C}$ \\
        Gyroscope bias & $\Vector{I}{b_{\bm{\omega}}}{}$ & $\Vector{}{b}{}$ \\
        Angular velocity measurement & $\Vector{I}{w}{}$ & $\Vector{}{w}{}$ \\
        Direction vector & $\Vector{G}{d}{}$ & $\Vector{}{d}{}$ \\
        \specialrule{.1em}{.1em}{.1em} 
    \end{tabular}
    \label{tab:conversion}
\end{table}

%% file: sections_supp/preliminaries.tex
\section{Mathematical Preliminaries}\label{sec:pre}

\subsection{Smooth Manifold}

Let $\xi \in \calM$ be an element of a smooth manifold $\calM$, $\tT\calM$ denotes the tangent bundle, $\tT_{\xi}\calM$ denotes the tangent space of $\calM$ at $\xi$ and ${\mathfrak{X}\left(\calM\right)}$ denotes the set of all smooth vector fields on $\calM$.

\subsection{Fréchet Derivative}

Let ${h \AtoB{\calM}{\calN}}$ be a continuous and differentiable map between manifolds. 
The differential is writen ${\td h \AtoB{\tT\calM}{\tT\calN}}$. Given ${\xi \in \calM}$ and ${\eta_{\xi} \in \tT_{\xi}\calM}$, the differential $\td h$ is evaluated pointwise by the Fréchet derivative\cit{Coleman2012CalculusSpaces} as
\begin{equation*}
    \td h\left(\xi\right)\left[\eta_{\xi}\right] = \Fr{\zeta}{\xi}h\left(\zeta\right)\left[\eta_{\xi}\right] \in \tT_{h\left(\xi\right)}\calN .
\end{equation*}

\subsection{Lie Group}

A Lie group $\grpG$ is a smooth manifold endowed with a smooth group multiplication that satisfies the group axioms. For any $X, Y \in \grpG$, the group multiplication is denoted $XY$, the group inverse $X^{-1}$ and the identity element $I$. 
In this work we will focus our attention on a special case of Lie groups, called matrix Lie groups, which are those whose elements are invertible square matrices in $\R^{n \times n}$ and where the group operation is the matrix multiplication.

\subsection{Lie Algebra}

For a given Lie group $\grpG$, the Lie algebra $\gothg$ is a vector space corresponding to $\tT_{I}\grpG$, together with a bilinear non-associative map ${[\cdot, \cdot] \AtoB{\gothg}{\gothg}}$ called Lie bracket, defined by
\begin{equation*}
    \left[\eta, \kappa \right] = \eta \kappa - \kappa \eta ,
\end{equation*}
for any $\eta, \kappa \in \gothg$.
For a matrix Lie-group $\gothg \subset \R^{n \times n}$ and the Lie-bracket is the matrix commutator. 

The Lie algebra $\gothg$ is isomorphic to a vector space $\R^{n}$ of dimension ${n =\mathrm{dim}\left(\gothg\right)}$.
Define the wedge map ${\left(\cdot\right)^{\wedge} \AtoB{\R^{n}}{\gothg}}$ and its inverse, the vee map ${\left(\cdot\right)^{\vee} \AtoB{\gothg}{\R^{n}}}$, as linear mappings between the vector space and the Lie algebra.

For any $X, Y \in \grpG$, left and right translation by $X$ are defined to be 
\begin{align*}
    &L_{X} \AtoB{\grpG}{\grpG}, \qquad L_{X}\left(Y\right) = XY ,\\
    &R_{X} \AtoB{\grpG}{\grpG}, \qquad R_{X}\left(Y\right) = YX. 
\end{align*} 
It follows that for any element of the tangent space $\eta_{Y} \in \tT_{Y}\grpG$, the differential of the left and right translations are
\begin{align*}
    &\td L_{X} \AtoB{\tT_{Y}\grpG}{\tT_{XY}\grpG}, \qquad \td L_{X}\left[\eta_{Y}\right] = X\eta_{Y} ,\\
    &\td R_{X} \AtoB{\tT_{Y}\grpG}{\tT_{YX}\grpG}, \qquad \td R_{X}\left[\eta_{Y}\right] = \eta_{Y}X .
\end{align*} 
where the expression $X n_Y$ and $n_Y X$ are group multiplication.

\subsection{Group Action and Homogeneous Space}

A right group action of a Lie group $\grpG$ on a differentiable manifold $\calM$ is a smooth map ${\phi \AtoB{\grpG\times \calM}{\calM}}$ that satisfies ${\phi\left(I, \xi\right) = \xi}$ and ${\phi\left(X, \phi\left(Y, \xi\right)\right) = \phi\left(YX, \xi\right)}$ for any $X,Y \in \grpG$ and $\xi \in \calM$.
A right group action $\phi$ induces a family of diffeomorphism ${\phi_X \AtoB{\calM}{\calM}}$ and smooth nonlinear projections ${\phi_{\xi} \AtoB{\grpG}{\calM}}$. The group action $\phi$ is said to be transitive if the induced projection $\phi_{\xi}$ is surjective, and then, in this case, $\calM$ is termed a homogeneous space.
A right group action $\phi$ induces a right group action on $\mathfrak{X}\left(\calM\right)$, ${\Phi \AtoB{\grpG \times \mathfrak{X}\left(\calM\right)}{\mathfrak{X}\left(\calM\right)}}$ defined by the push forward\cit{Mahony2021ObserverEquivariance} such that
\begin{equation*}
\Phi\left(X,f\right) = \td\phi_{X}f \circ \phi_{X^{-1}},
\end{equation*}
for each $f \in \mathfrak{X}\left(\calM\right)$, $\phi$-invariant vector field and $X \in \grpG$.

\subsection{The Semi-direct Product Group}

A semi-direct product group $\mathbf{G} \ltimes \mathbf{H}$ can be seen as a generalization of the direct product group $\grpG \times \grpH$ where the underlying set is given by the cartesian product of two groups $\grpG$, and $\grpH$ but, contrary to the direct product, in the semi-direct product, the group multiplication is defined with the group $\grpG$ that acts on a group $\grpH$ by a group automorphism. 
Note that the semi-direct product group $\mathbf{G} \ltimes \mathbf{H}$ with the trivial automorphism corresponds to the direct product group $\grpG \times \grpH$. In this work we will consider a semi-direct product symmetry group $\sdpgrpG := \mathbf{G} \ltimes \gothg$ where $\gothg$ is the Lie algebra of $\grpG$, a vector space group under addition. 
Let $A,B \in \grpG$ and $a,b \in \gothg$ and define $X = \left(A, a\right)$ and $Y = \left(B, b\right)$ elements of the symmetry group $\sdpgrpG$. 
Group multiplication is defined to be the semi-direct product ${YX = \left(BA, b + \Adsym{B}{a}\right)}$. 
The inverse element is ${X^{-1} = \left(A^{-1}, -\Adsym{A^{-1}}{a}\right)}$ and  identity element is ${\left(I, 0\right)}$. 


\subsection{Left Translation on the Semi-Direct Product Group}

Define the left translation on the semi-direct product group $\sdpgrpG$ by ${L_X : \sdpgrpG \to \sdpgrpG}$, ${L_X Y := XY}$.
Define a map ${\td L_X : \text{T}_Y \sdpgrpG \to \text{T}_{XY} \sdpgrpG}$ by
\begin{align*}
\td L_{(A,\, a)}[\eta_1,\, \eta_2] = (A \eta_1,\, \Ad_A [\eta_2]).
\end{align*}
\begin{lemma} \label{lem:dL}
$\td L_{(A,\, a)}$ is the differential of the left translation $L_{(A,\, a)}$.
\end{lemma}
\begin{proof}
Computing the differential of the left translation
\begin{align*}
\tD_{(B,\, b)}\Big|_{\left(I,\, 0\right)} &\left(\left(A,\, a\right)\left(B,\, b\right)\right)\left[\eta_1,\, \eta_2\right] \\
&= \tD_{\left(B,\, b\right)}\Big|_{\left(I,\, 0\right)} \left(AB,\, a + \Adsym{A}{b}\right)\left[\eta_1,\, \eta_2\right] \\
&= \left(A \eta_1,\, \Adsym{A}{\eta_2}\right) \\
&= \td L_{\left(A,\, a\right)}\left[\eta_1,\, \eta_2\right].
\end{align*}
\end{proof}

\subsection{Right Translation on the Semi-Direct Product Group}

Define the right translation on the semi-direct product group $\sdpgrpG$ by $R_X : \sdpgrpG \to \sdpgrpG$, $R_X Y := YX$.
Define a map $\td R_X : \text{T}_Y \sdpgrpG \to \text{T}_{YX} \sdpgrpG $ by
\begin{align*}
\td R_{\left(A,\, a\right)}\left[\eta_1,\, \eta_2\right] = \left(\eta_1 A,\, \eta_2 + \adsym{\eta_1}{a}\right).
\end{align*}
\begin{lemma} \label{lem:dR}
$\td R_{\left(A,\, a\right)}$ is the differential of the right translation $R_{\left(A,\, a\right)}$.
\end{lemma}
\begin{proof}
Computing the differential of the right translation
\begin{align*}
\tD_{\left(B,\, b\right)}\Big|_{\left(I,\, 0\right)} &(\left(B,\, b\right)\left(A,\, a\right))\left[\eta_1,\, \eta_2\right] \\
&= \tD_{\left(B,\, b\right)}\Big|_{\left(I,\, 0\right)} \left(BA,\, b + \Adsym{B}{a}\right)\left[\eta_1,\, \eta_2\right] \\
&= \left(\eta_1 A,\, \eta_2 + \adsym{\eta_1}{A}\right) \\
&= \td R_{\left(A,\, a\right)}\left[\eta_1,\, \eta_2\right].
\end{align*}
\end{proof}

%% file: sections_supp/eqf.tex
\section{Equivairant Filter Algorithm}
Let us recall the major components of an EqF for the biased attitude estimation with online calibration problem.

\subsection*{EqF Matrices}
Let ${\hat{X} = \left(\left(\hat{A},\,\hat{a}\right),\,\hat{B}\right) \in \grpG}$ be the \acl{eqf} state, with initial condition ${\hat{X}\left(0\right) = \left(\left(\eye,\,\mathbf{0}^{\wedge}\right),\,\eye\right)}$, the identity element of the symmetry group $\grpG$. 
The linearized error state matrix ${\mathbf{A}_{t}^{0}}$ (more details on how the state matrix ${\mathbf{A}_{t}^{0}}$ is derived can be found in \secref{A}) and the linearized output matrix ${\mathbf{C}^{0}}$ are defined as follows
\begin{subequations}
    \begin{minipage}{0.495\linewidth}
        \begin{equation*}
            \mathbf{A}_{t}^{0} = \begin{bmatrix}
                \mathbf{0} & -\mathbf{I} & \mathbf{0}\\
                \mathbf{0} & \Vector{}{\bm{\omega}}{0}^{\wedge} & \mathbf{0}\\
                \mathbf{0} & \mathbf{0} & \Vector{}{\bm{\omega}}{0}^{\wedge}
            \end{bmatrix} , \label{eq:At0_bas}
        \end{equation*}
    \end{minipage}
    \begin{minipage}{0.495\linewidth}
        \begin{equation*}
            \mathbf{C}^{0} = \begin{bmatrix}
                \Vector{}{d}{1}^{\wedge} & \mathbf{0} & \Vector{}{d}{1}^{\wedge}\\
                \Vector{}{d}{2}^{\wedge} & \mathbf{0} & \mathbf{0}
            \end{bmatrix}. \label{eq:C0_bas}
        \end{equation*}
    \end{minipage}
    \vspace{5pt}
\end{subequations}

The \ac{eqf} state-transition matrix ${\mathbf{\Phi}\left(t+\Delta t, t\right) = \mathbf{\Phi}}$ can be written as
\begin{equation*}\label{eq:stm}
    \mathbf{\Phi} = \begin{bmatrix}
    \eye & \mathbf{\Phi}_{12} & \mathbf{0}\\
    \mathbf{0} & \mathbf{\Phi}_{22} & \mathbf{0}\\
    \mathbf{0} & \mathbf{0} & \mathbf{\Phi}_{22}
    \end{bmatrix}
\end{equation*}
with
\begin{align*}
    \mathbf{\Phi}_{12} &= -\left(\Delta t \eye + \mathbf{\Psi}_{1}\Vector{}{\bm{\omega}}{0}^{\wedge} + \mathbf{\Psi}_{2} \Vector{}{\bm{\omega}}{0}^{\wedge}\Vector{}{\bm{\omega}}{0}^{\wedge}\right),\\
    \mathbf{\Phi}_{22} &= \eye + \mathbf{\Psi}_{3}\Vector{}{\bm{\omega}}{0}^{\wedge} + \mathbf{\Psi}_{1}\Vector{}{\bm{\omega}}{0}^{\wedge}\Vector{}{\bm{\omega}}{0}^{\wedge},\\
    \mathbf{\Psi}_{1} &= \frac{1-\cos\left(\norm{\Vector{}{\bm{\omega}}{0}} \Delta t\right)}{\norm{\Vector{}{\bm{\omega}}{0}}^2},\\
    \mathbf{\Psi}_{2} &= \frac{\norm{\Vector{}{\bm{\omega}}{0}} \Delta t - \sin\left(\norm{\Vector{}{\bm{\omega}}{0}} \Delta t{1}\right)}{\norm{\Vector{}{\bm{\omega}}{0}}^3},\\
    \mathbf{\Psi}_{3} &= \frac{\sin\left(\norm{\Vector{}{\bm{\omega}}{0}} \Delta t\right)}{\norm{\Vector{}{\bm{\omega}}{0}}}.
\end{align*}

For practical implementation aspects, it is useful to derive the state transition matrix $\mathbf{\Phi}$, defined in Equ. (18) in the limit of small angular velocity. therefore, taking the limit ${\lim_{\norm{\Vector{}{\bm{\omega}}{0}}\to0} \mathbf{\Phi}}$ and applying L'H\^{o}pital's rule yields
\begin{align*}
    \mathbf{\Phi}_{12} &\simeq -\Delta T\left(\eye + \frac{\Delta T}{2}\Vector{}{\bm{\omega}}{0}^{\wedge} + \frac{\Delta T^2}{6}\Vector{}{\bm{\omega}}{0}^{\wedge}\Vector{}{\bm{\omega}}{0}^{\wedge}\right),\\
    \mathbf{\Phi}_{22} &\simeq \eye + \Delta T\Vector{}{\bm{\omega}}{0}^{\wedge} + \frac{\Delta T^2}{2}\Vector{}{\bm{\omega}}{0}^{\wedge}\Vector{}{\bm{\omega}}{0}^{\wedge}.
\end{align*}

Let $ \mathbf{B}_{t}^{0}$, and $\mathbf{D}_{t}^{0}$ be the following matrices
\begin{subequations}
    \begin{minipage}{0.45\linewidth}
        \begin{equation*}
            \mathbf{B}_{t}^{0} = \begin{bmatrix}
                \hat{A} & \mathbf{0} & \mathbf{0}\\
                \mathbf{0} & \hat{A} & \mathbf{0}\\
                \mathbf{0} & \mathbf{0} & \hat{B}
            \end{bmatrix}
        \end{equation*}
    \end{minipage}
    \begin{minipage}{0.45\linewidth}
        \begin{equation*}
            \mathbf{D}_{t}^{0} = \begin{bmatrix}
                \hat{B} & \mathbf{0}\\
                \mathbf{0} & \hat{A}
            \end{bmatrix}
        \end{equation*}
    \end{minipage}
    \vspace{2.5pt}
\end{subequations}

Therefore, the detailed implementation of the proposed EqF algorithm is shown in \algref{eqf}. 

\subsection{\algref{eqf} Breakdown}
Let us explain in more detail some algorithmic choices. First of all, the EqF state ${\hat{X}}$ is implemented as two matrices. $\left(\hat{A},\, \hat{a}\right)$ as a $4 \times 4$ matrix
\begin{equation*}
    \left(\hat{A},\, \hat{a}\right) = \begin{bmatrix}
        \hat{A} & \hat{a}^{\vee}\\
        \mathbf{0} & 1
    \end{bmatrix} ,
\end{equation*}
and $\hat{B}$ as a $3 \times 3$ matrix. Similarly, $\Lambda_1$ is implemented as the following $4 \times 4$ matrix
\begin{equation*}
    \Lambda_1 = \begin{bmatrix}
        \left(\hat{A}^T\Vector{}{\omega}{0}\right)^{\wedge} & \Vector{}{\omega}{}^{\wedge}\hat{A}^T\hat{a}^{\vee}\\
        \mathbf{0} & 0
    \end{bmatrix} ,
\end{equation*}
while ${\Lambda_2}$ is clearly a $3 \times 3$ matrix. The state propagation is implemented as follows
\begin{align*}
    \begin{bmatrix}
        \hat{A} & \hat{a}^{\vee}\\
        \mathbf{0} & 1
    \end{bmatrix} &\gets 
    \begin{bmatrix}
        \hat{A} & \hat{a}^{\vee}\\
        \mathbf{0} & 1
    \end{bmatrix}\exp\left(\begin{bmatrix}
        \left(\hat{A}^T\Vector{}{\omega}{0}\right)^{\wedge} & \Vector{}{\omega}{}^{\wedge}\hat{A}^T\hat{a}^{\vee}\\
        \mathbf{0} & 0
    \end{bmatrix}\Delta t
    \right) ,\\
    \hat{B} &\gets \hat{B}\exp\left(\hat{B}^T\hat{A}\left(\hat{A}^T\Vector{}{\omega}{0}\right)^{\wedge}\hat{A}^T\hat{B}\Delta t\right) ,
\end{align*}
where ${\exp}$ is the matrix exponential. 

For what concerns the update part, the scaled equivariant innovation is a 9-vector ${\Vector{}{r}{} \in \R^{9}}$, therefore the notation ${\Vector{}{r}{i:k}}$ refers to the sub-vector of ${\Vector{}{r}{}}$ with starting index $i$, and ending index $k$. Similar to the propagation, the state update is implemented as follows
\begin{align*}
    \begin{bmatrix}
        \hat{A} & \hat{a}^{\vee}\\
        \mathbf{0} & 1
    \end{bmatrix} &\gets 
    \exp\left(\begin{bmatrix}
        \Vector{}{r}{1:3} & -\Vector{}{r}{4:6}\\
        \mathbf{0} & 0
    \end{bmatrix}\right)
    \begin{bmatrix}
        \hat{A} & \hat{a}^{\vee}\\
        \mathbf{0} & 1
    \end{bmatrix} ,\\
    \hat{B} &\gets \exp\left(\left(\Vector{}{r}{7:9}+\Vector{}{r}{1:3}\right)^{\wedge}\right)\hat{B} ,
\end{align*}

\SetInd{0.5em}{0.75em}
\newcommand\mycommfont[1]{\footnotesize\ttfamily{#1}}
\SetCommentSty{mycommfont}
\SetKwComment{Comment}{/* }{ */}
\SetKwProg{Loop}{loop}{}{end}
\SetKwProg{Propagation}{propagation}{}{end}
\SetKwProg{Update}{update}{}{end}
\begin{algorithm}[!htp]
\setstretch{1.15}
\caption{Implemented EqF}\label{alg:eqf}
\KwIn{Initial state: ${\hat{X} = \left(\left(\hat{A},\, \hat{a}\right)),\, \hat{B}\right) = \left(\left(\eye,\,\mathbf{0}^{\wedge}\right)\,\eye\right)}$}
\KwIn{Initial covariance: ${\bm{\Sigma}}$}
\KwIn{Known magnetic direction: ${\Vector{}{d}{1}}$}
\KwIn{Known measurement: ${\Vector{}{y}{2} = \left[0\,1\,0\right]^T}$}
\KwIn{Gyro measurement noise covariance: ${\bm{\Sigma}_{\omega}}$}
\KwIn{Random walk noise covariance: ${\bm{\Sigma}_{b}}$}
\KwIn{Direction measurements noise covariance: ${\bm{\Sigma}_{y}}$}
\Loop{}{
\KwIn{Gyro measurement: ${\bm{\omega}}$}
\Propagation{}{
$\Vector{}{\omega}{0} \coloneqq \hat{A}\Vector{}{\omega}{} + \hat{a}^{\vee}$\;
$\mathbf{\Phi}_{12} = -\Delta T\left(\eye + \frac{\Delta T}{2}\Vector{}{\bm{\omega}}{0}^{\wedge} + \frac{\Delta T^2}{6}\Vector{}{\bm{\omega}}{0}^{\wedge}\Vector{}{\bm{\omega}}{0}^{\wedge}\right)$\;
$\mathbf{\Phi}_{22} = \eye + \Delta T\Vector{}{\bm{\omega}}{0}^{\wedge} + \frac{\Delta T^2}{2}\Vector{}{\bm{\omega}}{0}^{\wedge}\Vector{}{\bm{\omega}}{0}^{\wedge}$\;
$\mathbf{\Phi} = \begin{bsmallmatrix}
    \eye & \mathbf{\Phi}_{12} & \mathbf{0}\\
    \mathbf{0} & \mathbf{\Phi}_{22} & \mathbf{0}\\
    \mathbf{0} & \mathbf{0} & \mathbf{\Phi}_{22}
\end{bsmallmatrix}$\;
$\mathbf{B}_{t}^{0} = \begin{bsmallmatrix}
    \hat{A} & \mathbf{0} & \mathbf{0}\\
    \mathbf{0} & \hat{A} & \mathbf{0}\\
    \mathbf{0} & \mathbf{0} & \hat{B}
\end{bsmallmatrix}$\;
$\bm{\Sigma}_{u} = \begin{bsmallmatrix}
    \bm{\Sigma}_{\omega} & \mathbf{0} & \mathbf{0}\\
    \mathbf{0} & \bm{\Sigma}_{b} & \mathbf{0}\\
    \mathbf{0} & \mathbf{0} & \mathbf{0}
\end{bsmallmatrix}$\;
${\bm{M}_c = \mathbf{B}_{t}^{0}\bm{\Sigma}_{u}{\mathbf{B}_{t}^{0}}^T}$\;
$\begin{bsmallmatrix}
    \hat{A} & \hat{a}^{\vee}\\
    \mathbf{0} & 1
\end{bsmallmatrix} \gets 
\begin{bsmallmatrix}
    \hat{A} & \hat{a}^{\vee}\\
    \mathbf{0} & 1
\end{bsmallmatrix}\exp\left(\begin{bsmallmatrix}
    \left(\hat{A}^T\Vector{}{\omega}{0}\right)^{\wedge} & \Vector{}{\omega}{}^{\wedge}\hat{A}^T\hat{a}^{\vee}\\
    \mathbf{0} & 0
\end{bsmallmatrix}\Delta t\right)$\;
$\hat{B} \gets \hat{B}\exp\left(\hat{B}^T\hat{A}\left(\hat{A}^T\Vector{}{\omega}{0}\right)^{\wedge}\hat{A}^T\hat{B}\Delta t\right)$\;
$\bm{\Sigma} \gets \mathbf{\Phi}\bm{\Sigma}\mathbf{\Phi}^T + \bm{M}_{c}\Delta t$\;
}
\KwIn{Magnetometer measurements: ${\mathbf{m}}$}
\KwIn{GNSS position measurements: ${\mathbf{p}_1,\,\mathbf{p}_2}$}
\Update{}{
${\Vector{}{y}{1} = \frac{\mathbf{m}}{\norm{\mathbf{m}}}}$\;
${\Vector{}{d}{2} = \frac{\mathbf{p}_1-\mathbf{p}_2}{\norm{\mathbf{p}_1-\mathbf{p}_2}}}$\;
$\mathbf{C}^{0} = \begin{bsmallmatrix}
\Vector{}{d}{1}^{\wedge} & \mathbf{0} & \Vector{}{d}{1}^{\wedge}\\
\Vector{}{d}{2}^{\wedge} & \mathbf{0} & \mathbf{0}
\end{bsmallmatrix}$\;
$\mathbf{D}_{t}^{0} = \begin{bsmallmatrix}
    \hat{B} & \mathbf{0}\\
    \mathbf{0} & \hat{A}
\end{bsmallmatrix}$\;
$\bm{N} = \mathbf{D}_{t}^{0}\bm{\Sigma}_{y}{\mathbf{D}_{t}^{0}}^T$\;
$\mathbf{S} = \mathbf{B}^{0}\bm{\Sigma}{\mathbf{C}^{0}}^T + \bm{N}$\;
$\mathbf{K} = \bm{\Sigma}{\mathbf{C}^{0}}^T\mathbf{S}^{-1}$\;
$\Vector{}{r}{} = \mathbf{K}\begin{bmatrix}
    \hat{B}\Vector{}{y}{1}\\
    \hat{A}\Vector{}{y}{2}
\end{bmatrix}$\;
$\begin{bsmallmatrix}
    \hat{A} & \hat{a}^{\vee}\\
    \mathbf{0} & 1
\end{bsmallmatrix} \gets 
\exp\left(\begin{bsmallmatrix}
    \Vector{}{r}{1:3}^{\wedge} & -\Vector{}{r}{4:6}\\
    \mathbf{0} & 0
\end{bsmallmatrix}\right)
\begin{bsmallmatrix}
    \hat{A} & \hat{a}^{\vee}\\
    \mathbf{0} & 1
\end{bsmallmatrix}$\;
$\hat{B} \gets \exp\left(\left(\Vector{}{r}{7:9}+\Vector{}{r}{1:3}\right)^{\wedge}\right)\hat{B}$\;
$\bm{\Sigma} \gets \left(\eye - \mathbf{K}\mathbf{C}^0\right)\bm{\Sigma}$\;
}
}
\end{algorithm}

%% file: sections_supp/iekf.tex
\section{State-of-the-art (Imperfect-IEKF) Design}

This section quickly summarizes previous work\cit{bonnabel2007left},\cit{7523335},\cit{barrau:tel-01247723}, \cite{doi:10.1177/0278364919894385},\cit{Cohen2020NavigationEstimation}, on the \ac{iekf}, and, following that work, a novel discrete-time formulation of the Imperfect-\ac{iekf} for the \acl{bas} with $N = 2$ direction measurements and $n = 1$ extrinsic calibration state is derived.

\subsection{IEKF Filter Matrices}
Let ${\xi = \left(\Rot{G}{I},\, \Vector{I}{b}{\bm{\omega}},\, \Rot{I}{S}\right) \in \SO\left(3\right) \times \R^{3} \times \SO\left(3\right)}$ be the system state, we define the error ${\bm{\eta} = \left(\bm{\eta}_{\Rot{}{}},\, \bm{\eta}_{\Vector{}{b}{}},\, \bm{\eta}_{\mathbf{C}}\right)}$ to be right-invariant for the attitude and calibration states, but Euclidean for the biases, therefore
\begin{equation*}
    \bm{\eta} = \left(\bm{\eta}_{\Rot{}{}}^{\wedge},\, \bm{\eta}_{\Vector{}{b}{}},\, \bm{\eta}_{\mathbf{C}}^{\wedge}\right) = \left(\Rot{G}{I}\hatRot{G}{I}^T,\, \Vector{I}{b}{\bm{\omega}} - \hatVector{I}{b}{\bm{\omega}},\, \Rot{I}{S}\hatRot{I}{S}^T\right) .
\end{equation*}
For the sake of clarity, we will omit superscript and subscript from the derivation that follows. The linearized error dynamics for the \acs{iekf} can then be computed by differentiating the defined error $\bm{\eta}$, therefore
\begin{align*}
    \begin{split}
        &\dot{\bm{\eta}_{\mathbf{R}}^{\wedge}} = \dot{\mathbf{R}}\hat{\mathbf{R}}^{T} + \mathbf{R}\dot{\hat{\mathbf{R}}}^{T} = \mathbf{R}\left(\bm{\omega} - \mathbf{b_\omega}\right)^{\wedge}\hat{\mathbf{R}}^{T} - \mathbf{R}\hat{\mathbf{R}}^{T}\dot{\hat{\mathbf{R}}}\hat{\mathbf{R}}^{T}\\
        & \quad\; = \mathbf{R}\left(\bm{\omega} - \mathbf{b_\omega}\right)^{\wedge}\hat{\mathbf{R}}^{T} - \mathbf{R}\left(\bm{\omega}- \hat{\mathbf{b_\omega}}\right)^{\wedge}\hat{\mathbf{R}}^{T}\\
        & \quad\; = -\mathbf{R}\left(\mathbf{b_\omega}- \hat{\mathbf{b_\omega}}\right)^{\wedge}\hat{\mathbf{R}}^{T}\\
        & \quad\; \approx -\left(\mathbf{I} + \bm{\eta}_{\mathbf{R}}^{\wedge}\right)\hat{\mathbf{R}}\left(\bm{\eta}_{\mathbf{b}}\right)^{\wedge}\hat{\mathbf{R}}^{T} \approx \left(-\hat{\mathbf{R}}\bm{\eta}_{\mathbf{b}}\right)^{\wedge} ,
    \end{split}\\
    &\dot{\bm{\eta}_{\Vector{}{b}{}}} = \prescript{}{}{\dot{\bm{b}}}_{\omega} - \prescript{}{}{\dot{\hat{\bm{b}}}}_{\omega} = \mathbf{0} ,\\
    &\dot{\bm{\eta}_{\mathbf{C}}^{\wedge}} = \dot{\mathbf{C}}\hat{\mathbf{C}}^{T} + \mathbf{C}\dot{\hat{\mathbf{C}}}^{T} = \mathbf{0}^{\wedge} .
\end{align*}
We can then write
\begin{equation*}
    \dot{\bm{\eta}} = \mathbf{F}\bm{\eta} , \qquad\text{with}\qquad
    \mathbf{F} = \begin{bmatrix}
    \mathbf{0} & -\hatRot{}{} & \mathbf{0}\\
    \mathbf{0} & \mathbf{0} & \mathbf{0}\\
    \mathbf{0} & \mathbf{0} & \mathbf{0}
    \end{bmatrix} .
\end{equation*}

\subsection{IEKF Formulation and Practical Implementation}
\label{sec:iekf_implementation}

Let ${\bm{\Sigma} \in \PD\left(9\right) \subset \R^{9 \times 9}}$ be the error covariance matrix, with initial condition ${\bm{\Sigma}\left(0\right) = \bm{\Sigma}_0}$. Let ${\bm{M}_c \in \PD\left(9\right) \subset \R^{9 \times 9}}$ be the continuous-time input gain matrix. The propagation equations of the \acs{iekf} write
\begin{align*}
    &\dot{\xi} = f_{\bm{\omega}}\left(\xi\right) ,\\
    &\dot{\bm{\Sigma}} = \mathbf{F}\bm{\Sigma} + \bm{\Sigma}\mathbf{F}^T + \bm{M}_c .\label{eq:iekf_covariance_ode}
\end{align*}

A discrete-time implementation of the filter's propagation phase requires integration of the aforementioned equations in the time-step $\Delta T$ in between gyro measurements. The \acs{iekf} state-transition matrix ${\mathbf{\Phi}\left(t+\Delta t, t\right) = \mathbf{\Phi}}$ can be written
\begin{equation*}
    \mathbf{\Phi} = \eye + \mathbf{F}\Delta T\in \R^{9\times 9}.
\end{equation*}

This is not an approximation since ${\mathbf{F}^k = \mathbf{0} \; \forall \; k>1}$.
Therefore, the discrete-time implementation of the \acs{iekf}'s propagation phase follows
\begin{align*}
    \hat{\xi}_{k+1}^{-} &= \hat{\xi}_{k}^{+} + \int_{0}^{\Delta T} f_{\bm{\omega}\left(t\right)}\left(\xi\left(t\right)\right)\; dt,\\
    \bm{\Sigma}_{k+1}^{-} &= \mathbf{\Phi}\bm{\Sigma}_{k}^{+}\mathbf{\Phi}^T + \bm{M}_d,
\end{align*}
where $\bm{M}_d$ can be approximated\cit{Sola2017QuaternionFilter} by ${\bm{M}_d \simeq \bm{M}_c\Delta T}$.

Let the output map be the one defined in Equ. (3), with $n = 1$, and $N = 2$ as before. Therefore, the output residual, and its linearization with respect to the error $\bm{\eta}$ write
\begin{align*}
    \left(\hatRot{}{}\hat{\mathbf{C}}\Vector{}{y}{1},\,\hatRot{}{}\Vector{}{y}{2}\right) &= \left(\hatRot{}{}\hat{\mathbf{C}}\mathbf{C}^T\Rot{}{}^T\Vector{}{d}{1},\,\hatRot{}{}\Rot{}{}^T\Vector{}{d}{2}\right)\\
    &= \left(\hatRot{}{}\hat{\mathbf{C}}\mathbf{C}^T\hatRot{}{}^T\hatRot{}{}\Rot{}{}^T\Vector{}{d}{1},\,\hatRot{}{}\Rot{}{}^T\Vector{}{d}{2}\right)\\
    &\simeq \left(\Vector{}{d}{1}^{\wedge}\bm{\eta}_{\mathbf{R}} + \Vector{}{d}{1}^{\wedge}\hatRot{}{}\bm{\eta}_{\mathbf{C}},\,\Vector{}{d}{2}^{\wedge}\bm{\eta}_{\mathbf{R}}\right).
\end{align*}

Given the output residual and its linearization, we derive the \acs{iekf} update equations, in particular let ${\bm{N} \in \PD\left(6\right) \subset \R^{6 \times 6}}$ be the output gain matrix, the \acl{iekf} update equations write
\begin{align*}
    &\mathbf{H} = \begin{bmatrix}
    \Vector{}{d}{1}^{\wedge} & \mathbf{0} & \Vector{}{d}{1}^{\wedge}\hatRot{}{}\\
    \Vector{}{d}{2}^{\wedge} & \mathbf{0} & \mathbf{0}
    \end{bmatrix} ,\\
    &\mathbf{S} = \mathbf{H}\bm{\Sigma}\mathbf{H}^T + \bm{N} ,\\
    &\mathbf{K} = \bm{\Sigma}\mathbf{H}^T\bm{\Sigma}^{-1} ,\\
    &\hat{\xi}_{k+1}^{+} = \exp\left(\left(\mathbf{K}\begin{bmatrix}
        \hatRot{}{}\hat{\mathbf{C}}\Vector{}{y}{1} - \Vector{}{d}{1}\\
        \hatRot{}{}\Vector{}{y}{2} - \Vector{}{d}{2}
    \end{bmatrix}\right)^{\wedge}\right)\hat{\xi}_{k+1}^{-} ,\\
    &\bm{\Sigma}_{k+1}^{+} = \left(\eye - \mathbf{K}\mathbf{H}\right)\bm{\Sigma}_{k+1}^{-}.
\end{align*}

The \acs{iekf} needs to deal with noise in the measured input variables and in the output measurements, and hence the gain matrices $\bm{M}_c$ and $\bm{N}$ need to be defined accordingly to the covariance matrices of the measurements noise. Let ${\bm{\omega}_m = \bm{\omega} + \bm{\omega}_{\mathbf{n}}}$ be the measured input, where ${\mathbf{n}}$ is additive white Gaussian noise with covariance ${\bm{\Sigma}_{\omega}}$, therefore, the \acs{iekf} linearized error dynamics in the attitude change accordingly to
\begin{align*}
    &\dot{\bm{\eta}_{\mathbf{R}}} = \dot{\mathbf{R}}\hat{\mathbf{R}}^{T} + \mathbf{R}\dot{\hat{\mathbf{R}}}^{T} = \mathbf{R}\left(\bm{\omega}_m - \bm{\omega}_{\mathbf{n}} - \mathbf{b_\omega}\right)^{\wedge}\hat{\mathbf{R}}^{T} - \mathbf{R}\hat{\mathbf{R}}^{T}\dot{\hat{\mathbf{R}}}\hat{\mathbf{R}}^{T}\\
    & \quad\; = \mathbf{R}\left(\bm{\omega}_m - \bm{\omega}_{\mathbf{n}} - \mathbf{b_\omega}\right)^{\wedge}\hat{\mathbf{R}}^{T} - \mathbf{R}\left(\bm{\omega}_m - \hat{\mathbf{b_\omega}}\right)^{\wedge}\hat{\mathbf{R}}^{T}\\
    & \quad\; = -\mathbf{R}\left(\mathbf{b_\omega}- \hat{\mathbf{b_\omega}}\right)^{\wedge}\hat{\mathbf{R}}^{T} - \mathbf{R}\left(\bm{\omega}_{\mathbf{n}}\right)^{\wedge}\hat{\mathbf{R}}^{T}\\
    & \quad\; \approx -\left(\mathbf{I} + \bm{\eta}_{\mathbf{R}}^{\wedge}\right)\left(\hat{\mathbf{R}}\left(\bm{\eta}_{\mathbf{b}}\right)^{\wedge}\hat{\mathbf{R}}^{T} - \hatRot{}{}\bm{\omega}_{\mathbf{n}}\hatRot{}{}^T\right)^{\wedge}\\
    & \quad\;\approx \left(-\hat{\mathbf{R}}\bm{\eta}_{\mathbf{b}} + \hat{\mathbf{R}}\bm{\omega}_{\mathbf{n}}\right)^{\wedge} ,
\end{align*}
Similar considerations hold for the gyro bias, defined as a random walk process with additive white Gaussian noise $\bm{b}_{\mathbf{n}}$, with covariance ${\bm{\Sigma}_{b}}$. therefore
\begin{align*}
    &\dot{\bm{\eta}_{\Vector{}{b}{}}} = \bm{b}_{\mathbf{n}} ,
\end{align*}
therefore, we define
\begin{equation*}
\mathbf{B} = \begin{bmatrix}
    \hatRot{}{} & \mathbf{0} & \mathbf{0}\\
    \mathbf{0} & \eye & \mathbf{0}\\
    \mathbf{0} & \mathbf{0} & \hat{\mathbf{C}}
    \end{bmatrix}
\end{equation*}

Similarly, let $\Vector{}{y}{} + \mathbf{n}$ be a real-world direction measurement, where ${\mathbf{n}}$ is additive white Gaussian noise with covairance ${\bm{\Sigma}_{y}}$, then the output residual is written
\begin{align*}
    \left(\hatRot{}{}\hat{\mathbf{C}}\Vector{}{y}{1},\,\hatRot{}{}\Vector{}{y}{2}\right) + \left(\hatRot{}{}\hat{\mathbf{C}}\mathbf{n}_1,\,\hatRot{}{}\mathbf{n}_2\right) ,
\end{align*}
therefore, we define
\begin{equation*}
    \mathbf{D} = \begin{bmatrix}
    \hatRot{}{}\hat{\mathbf{C}} & \mathbf{0}\\
    \mathbf{0} & \hatRot{}{}
    \end{bmatrix}.
\end{equation*}

Let ${\bm{\Sigma}_{u}} = diag\left(\bm{\Sigma}_{\omega},\,\bm{\Sigma}_{b},\,\mathbf{0}\right)$ be the input measurement covariance, then the gain matrices $\bm{M}_c$ and $\bm{N}$ are then defined accordingly to
\begin{align*}
    &\bm{M}_c =  \mathbf{B}\bm{\Sigma}_{u}\mathbf{B}^T ,\\
    &\bm{N} =  \mathbf{D}\bm{\Sigma}_y\mathbf{D}^T .
\end{align*}

Finally, the implemented Imperfect-IEKF algorithm is detailed in \algref{iekf}.

\SetInd{0.5em}{0.75em}
\begin{algorithm}[!htp]
\setstretch{1.15}
\caption{Implemented Imperfect-IEKF}\label{alg:iekf}
\KwIn{Initial state: ${\hat{\xi} = \left(\hatRot{}{},\, \hatVector{}{b}{},\, \hat{\mathbf{C}}\right)}$}
\KwIn{Initial covariance: ${\bm{\Sigma}}$}
\KwIn{Known magnetic direction: ${\Vector{}{d}{1}}$}
\KwIn{Known measurement: ${\Vector{}{y}{2} = \left[0\,1\,0\right]^T}$}
\KwIn{Gyro measurement noise covariance: ${\bm{\Sigma}_{\omega}}$}
\KwIn{Random walk noise covariance: ${\bm{\Sigma}_{b}}$}
\KwIn{Direction measurements noise covariance: ${\bm{\Sigma}_{y}}$}
\Loop{}{
\KwIn{Gyro measurement: ${\bm{\omega}}$}
\Propagation{}{
$\mathbf{F} = \begin{bsmallmatrix}
    \mathbf{0} & -\hatRot{}{} & \mathbf{0}\\
    \mathbf{0} & \mathbf{0} & \mathbf{0}\\
    \mathbf{0} & \mathbf{0} & \mathbf{0}
\end{bsmallmatrix}$\;
$\mathbf{\Phi} = \mathbf{I} + \mathbf{F}\Delta t$\;
$\mathbf{B} = \begin{bsmallmatrix}
    \hatRot{}{} & \mathbf{0} & \mathbf{0}\\
    \mathbf{0} & \eye & \mathbf{0}\\
    \mathbf{0} & \mathbf{0} & \hat{\mathbf{C}}
\end{bsmallmatrix}$\;
$\bm{\Sigma}_{u} = 
\begin{bsmallmatrix}
    \bm{\Sigma}_{\omega} & \mathbf{0} & \mathbf{0}\\
    \mathbf{0} & \bm{\Sigma}_{b} & \mathbf{0}\\
    \mathbf{0} & \mathbf{0} & \mathbf{0}
\end{bsmallmatrix}$\;
$\bm{M}_c = \mathbf{B}\bm{\Sigma}_{u}\mathbf{B}^T$\;
$\hat{\xi} \gets \left(\hatRot{}{}\exp\left(\left(\bm{\omega} - \hatVector{}{b}{}\right)^{\wedge}\Delta t\right),\,  \hatVector{}{b}{},\, \hat{\mathbf{C}}\right)$\;
$\bm{\Sigma} \gets \mathbf{\Phi}\bm{\Sigma}\mathbf{\Phi}^T + \bm{M}_{c}\Delta t$\;
}
\KwIn{Magnetometer measurements: ${\mathbf{m}}$}
\KwIn{GNSS position measurements: ${\mathbf{p}_1,\,\mathbf{p}_2}$}
\Update{}{
${\Vector{}{y}{1} = \frac{\mathbf{m}}{\norm{\mathbf{m}}}}$\;
${\Vector{}{d}{2} = \frac{\mathbf{p}_1-\mathbf{p}_2}{\norm{\mathbf{p}_1-\mathbf{p}_2}}}$\;
$\mathbf{H} = \begin{bsmallmatrix}
\Vector{}{d}{1}^{\wedge} & \mathbf{0} & \Vector{}{d}{1}^{\wedge}\hatRot{}{}\\
\Vector{}{d}{2}^{\wedge} & \mathbf{0} & \mathbf{0}
\end{bsmallmatrix}$\;
$\mathbf{D} = \begin{bsmallmatrix}
    \hatRot{}{}\hat{\mathbf{C}} & \mathbf{0}\\
    \mathbf{0} & \hatRot{}{}
\end{bsmallmatrix}$\;
$\bm{N} = \mathbf{D}\bm{\Sigma}_{y}\mathbf{D}^T$\;
$\mathbf{S} = \mathbf{H}\bm{\Sigma}\mathbf{H}^T + \bm{N}$\;
$\mathbf{K} = \bm{\Sigma}\mathbf{H}^T{\bm{\Sigma}}^{-1}$\;
$\hat{\xi} \gets \exp\left(\left(\mathbf{K}
\begin{bsmallmatrix}
    \hatRot{}{}\hat{\mathbf{C}}\Vector{}{y}{1} - \Vector{}{d}{1}\\
    \hatRot{}{}\Vector{}{y}{2} - \Vector{}{d}{2}
\end{bsmallmatrix}
\right)^{\wedge}\right)\hat{\xi}$\;
$\bm{\Sigma} \gets \left(\eye - \mathbf{K}\mathbf{H}\right)\bm{\Sigma}$\;
}
}
\end{algorithm}

%% file: sections_supp/appendix.tex
\section{Extended Proofs and Appendix}

\subsection{Proof of Lemma 3.1}
\begin{manuallemma}{3.1}
Define ${\phi \AtoB{\grpG \times \calM}{\calM}}$ as
\begin{equation*}
    \phi\left(X, \xi\right) \coloneqq \left(\Rot{}{}A,\, A^T\left(\Vector{}{b}{} - a^{\vee}\right),\, A^T\mathbf{C}B\right) \in \calM .
\end{equation*}
Then, $\phi$ is a transitive right group action of $\grpG$ on $\calM$.
\end{manuallemma}

\begin{proof}
Let ${X, Y \in \grpG}$ and $\xi \in \calM$. Then, 
\begin{align*}
    &\phi\left(X,\phi\left(Y, \xi\right)\right)\\
    &= \phi\left(X,\, \left(\Rot{}{}A_Y,\, A_Y^T\left(\Vector{}{b}{} - a_Y^{\vee}\right),\, A_Y^T\mathbf{C}B_Y\right)\right)\\
    &= \left(\Rot{}{}A_YA_X,\, A_X^T\left(A_Y^T\left(\Vector{}{b}{} - a_Y^{\vee}\right) - a_X^{\vee}\right),\,A_X^TA_Y^T\mathbf{C}B_YB_X\right)\\
    &= \left(\Rot{}{}\left(A_YA_X\right),\, \left(A_YA_X\right)^T\left(\Vector{}{b}{} - \left(a_Y^{\vee} + A_Ya_X^{\vee}\right)\right),\right.\\
    &\quad\left.\left(A_YA_X\right)^T\mathbf{C}\left(B_YB_X\right)\right) = \phi\left(YX,\, \xi\right) .
\end{align*}
This shows that $\phi$ is a valid right group action. Then, ${\forall \; \xi_1, \xi_2 \in \calM}$ we can always write the group element ${Z = \left(\left(\Rot{}{}_{1}^{T}\Rot{}{}_{2},\, \Vector{}{b}{1} - \Rot{}{}_{1}^{T}\Rot{}{}_{2}\Vector{}{b}{2}\right),\,\mathbf{C}_1^T\Rot{}{}_{1}^{T}\Rot{}{}_{2}\mathbf{C}_2\right)}$, such that
\begin{align*}
    \phi\left(Z, \xi_1\right) &= \left(\left(\Rot{}{}_{1}\Rot{}{}_{1}^{T}\Rot{}{}_{2},\, \left(\Rot{}{}_{1}^{T}\Rot{}{}_{2}\right)^{T}\left(\Vector{}{b}{1} - \Vector{}{b}{1} + \Rot{}{}_{1}^{T}\Rot{}{}_{2}\Vector{}{b}{2}\right)\right),\right.\\
    &\quad\left.\left(\Rot{}{}_{1}^{T}\Rot{}{}_{2}\right)^{T}\mathbf{C}_1\mathbf{C}_1^T\Rot{}{}_{1}^{T}\Rot{}{}_{2}\mathbf{C}_2\right)\\
    &= \left(\left(\Rot{}{}_{2},\, \Vector{}{b}{2}\right),\,\mathbf{C}_2\right) = \xi_2 ,
\end{align*}
which demonstrates the transitive property of the group action.
\end{proof}

\subsection{Proof of Lemma 3.3}
\begin{manuallemma}{3.3}
Define ${\psi \AtoB{\grpG \times \vecL}{\vecL}}$ as
\begin{equation*}
    \begin{split}
        \psi\left(X,\mathbf{u}\right) &\coloneqq \left(A^T\left(\bm{\omega} - a^{\vee}\right),\, \mathbf{0},\, \mathbf{0}\right) ,
    \end{split}
\end{equation*}
Then, $\psi$ is a right group action of $\grpG$ on $\vecL$.
\end{manuallemma}

\begin{proof}
Let ${X, Y \in \grpG}$ and $\mathbf{u} \in \vecL$. Then, 
\begin{align*}
    &\psi\left(X,\psi\left(Y, \mathbf{u}\right)\right)\\
    &= \psi\left(X,\left(A_Y^T\left(\bm{\omega} - a_Y^{\vee}\right),\, \mathbf{0},\, \mathbf{0}\right)\right)\\
    &= \left(A_X^T\left(A_Y^T\left(\bm{\omega} - a_Y^{\vee}\right) - a_X^{\vee}\right),\, \mathbf{0},\, \mathbf{0}\right)\\
    &= \left(\left(A_YA_X\right)^T\left(\bm{\omega} - \left(a_Y^{\vee} + A_Ya_X^{\vee}\right)\right),\, \mathbf{0},\, \mathbf{0}\right)\\
    &= \psi\left(YX, \mathbf{u}\right) .
\end{align*}
Thus, proving that $\psi$ is a valid right group action.
\end{proof}

\subsection{Proof of Theorem 3.5}
\begin{manualtheorem}{3.5}
The \acl{bas} in Equ. (2) is equivariant under the actions $\phi$ Equ. (4) and $\psi$ in Equ. (5) of the symmetry group $\grpG$. That is
\begin{equation*}
    f_0\left(\xi\right) + f_{\psi_{X}\left(\bm{u}\right)}\left(\xi\right) = \Phi_{X}f_0\left(\xi\right) + \Phi_{X}f_{\bm{u}}\left(\xi\right) .
\end{equation*}
\end{manualtheorem}
\begin{proof}
Let ${X \in \grpG}$, ${\xi \in \calM}$ and ${\mathbf{u} \in \vecL}$, then the inverse of the group action writes
\begin{equation*}
    \phi\left(X^{-1},\,\xi\right) \coloneqq \left(\left(\Rot{}{}A^T,\, A\Vector{}{b}{} + a^{\vee}\right),\, A\mathbf{C}B^T\right) \in \calM
\end{equation*}
therefore
\begin{align*}
    &\Phi_{X}f_0\left(\xi\right) + \Phi_{X}f_{\bm{u}}\left(\xi\right) \\
    &= \left(\left(\Rot{}{}\left(A^T\left(\bm{\omega} - a^{\vee}\right)^{\wedge} - \Vector{}{b}{}^{\wedge}A^T\right)A,\, \mathbf{0}\right),\,\mathbf{0}^{\wedge}\right) \\
    &= \left(\left(\Rot{}{}\left(\left(A^T\left(\bm{\omega} - a^{\vee}\right)\right)^{\wedge} - \Vector{}{b}{}^{\wedge}\right),\, \mathbf{0}\right),\,\mathbf{0}^{\wedge}\right) \\
    &= f_0\left(\xi\right) + f_{\psi_{X}\left(\mathbf{u}\right)}\left(\xi\right) ,
\end{align*}
proving the equivariance of the system.
\end{proof}

\subsection{Proof of Lemma 3.6}
\begin{manuallemma}{3.6}
Define ${\rho \AtoB{\grpG \times \calN}{\calN}}$ as
\begin{equation*}
    \rho\left(X,\Vector{}{y}{}\right) \coloneqq \left(B^T\Vector{}{y}{1},\,A^T\Vector{}{y}{2}\right) .
\end{equation*}
Then, the configuration output defined in Equ. (3) is equivariant.
\end{manuallemma}

\begin{proof}
    Let ${X \in \grpG}$ and ${h\left(\xi\right) = \left(\mathbf{C}^T\Rot{}{}^T\Vector{}{d}{1},\,\Rot{}{}^T\Vector{}{d}{2}\right) \in \calN}$ then,
    \begin{equation*}
        \rho\left(X, h\left(\xi\right)\right) = \left(B^T\mathbf{C}^T\Rot{}{}^T\Vector{}{d}{1},\,A^T\Rot{}{}^T\Vector{}{d}{2}\right) = h\left(\phi\left(X,\xi\right)\right) .
    \end{equation*}
    This proves the output equivariance. 
\end{proof}

\subsection{Proof of Theorem 3.8}
\begin{manualtheorem}{3.8}
Define ${\Lambda \AtoB{\calM \times \vecL} \gothg}$ as
\begin{equation*}
    \Lambda\left(\xi, \mathbf{u}\right) \coloneqq \left(\left(\left(\Vector{}{\bm{\omega}}{}^{\wedge} - \Vector{}{b}{}^{\wedge}\right),\, - \Vector{}{\omega}{}^{\wedge}\Vector{}{b}{}\right),\, \mathbf{C}^T\left(\Vector{}{\bm{\omega}}{}^{\wedge} - \Vector{}{b}{}^{\wedge}\right)\mathbf{C}\right) .
\end{equation*}
Then, the map ${\Lambda\left(\xi, \mathbf{u}\right)}$ is an equivariant lift for the system in Equ. (2) with respect to the defined symmetry group.
\end{manualtheorem}
\begin{proof}
Let ${\xi \in \calM}, \mathbf{u} \in \vecL$, then
\begin{align*}
    &d\phi_{\xi}\left(\mathbf{I}\right)\left[\Lambda\left(\xi, \mathbf{u}\right)\right]\\
    &= \left(\left(\Rot{}{}\left(\Vector{}{\bm{\omega}}{}^{\wedge} - \Vector{}{b}{}^{\wedge}\right),\, - \left(\Vector{}{\bm{\omega}}{}^{\wedge} - \Vector{}{b}{}^{\wedge}\right)\Vector{}{b}{} - \left(- \Vector{}{\omega}{}^{\wedge}\Vector{}{b}{}\right)\right),\right.\\
    &\quad\left.\,\mathbf{C}\mathbf{C}^T\left(\Vector{}{\bm{\omega}}{}^{\wedge} - \Vector{}{b}{}^{\wedge}\right)\mathbf{C} - \left(\Vector{}{\bm{\omega}}{}^{\wedge} - \Vector{}{b}{}^{\wedge}\right)\mathbf{C}\right)\\
    &= \left(\left(\Rot{}{}\left(\Vector{}{\bm{\omega}}{}^{\wedge} - \Vector{}{b}{}^{\wedge}\right),\, \mathbf{0}\right),\, \mathbf{0}^{\wedge}\right) = f_0\left(\xi\right) + f_{\mathbf{u}}\left(\xi\right) .
\end{align*}
To demonstrate the equivariance of the lift we have to show that the condition ${\Adsym{X}{\Lambda\left(\phi_{X}\left(\xi\right),\psi_{X}\left(\mathbf{u}\right)\right)} = \Lambda\left(\xi, \mathbf{u}\right)}$ holds. Let ${\xi \in \calM}, \mathbf{u} \in \vecL$ and ${X \in \grpG}$, then
\begin{align*}
    &\Adsym{X}{\Lambda\left(\phi_{X}\left(\xi\right),\psi_{X}\left(\mathbf{u}\right)\right)} = X\Lambda\left(\phi_{X}\left(\xi\right),\psi_{X}\left(\mathbf{u}\right)\right)X^{-1}\\
    &= \left(\left(\left(AA^T\left(\Vector{}{\bm{\omega}}{} - \Vector{}{b}{}\right)\right)^{\wedge},\, -A\left(A^T\left(\Vector{}{\bm{\omega}}{} - a\right)\right)^{\wedge}A^T\left(\Vector{}{b}{} - a\right) -\right.\right.\\
    &\left.\left. - \left(AA^T\left(\Vector{}{\bm{\omega}}{} - \Vector{}{b}{}\right)\right)^{\wedge}a\right),\,B\left(B^T\mathbf{C}^T\left(\Vector{}{\bm{\omega}}{} - \Vector{}{b}{}\right)^{\wedge}\mathbf{C}B\right)B^T\right)\\
    &=\left(\left(\left(\Vector{}{\bm{\omega}}{} - \Vector{}{b}{}\right)^{\wedge},\,-\left(\Vector{}{\bm{\omega}}{} - a\right)^{\wedge}\left(\Vector{}{b}{} - a\right)-\left(\Vector{}{\bm{\omega}}{} - \Vector{}{b}{}\right)^{\wedge}a\right),\right.\\
    &\quad\left.\mathbf{C}^T\left(\Vector{}{\bm{\omega}}{} - \Vector{}{b}{}\right)^{\wedge}\mathbf{C}\right)\\
    &=\left(\left(\left(\Vector{}{\bm{\omega}}{} - \Vector{}{b}{}\right)^{\wedge},\,-\Vector{}{\bm{\omega}}{}^{\wedge}\Vector{}{b}{}\right),\,\mathbf{C}^T\left(\Vector{}{\bm{\omega}}{} - \Vector{}{b}{}\right)^{\wedge}\mathbf{C}\right) = \Lambda\left(\xi, \mathbf{u}\right) .
\end{align*}
\end{proof}

\subsection{Derivation of ${\mathbf{A}_{t}^{0}}$ in Equ. 11, and Equ. 14.a}\label{sec:A}
The state error in the homogeneous space is defined to be ${e = \phi_{\hat{X}^{-1}}\left(\xi\right) \in \calM}$. Therefore, to compute the linearized error dynamics we first need to choose an origin $\xi_0$. The origin is chosen to be the identity of the homogeneous space, ${\xi_0 = \text{id}}$.
Then, we need to select local coordinates on the homogeneous space, to do so, we need to define a chart and a chart transition map ${\vartheta \AtoB{\mathcal{U}_{\xi_0} \subset \calM}{\R^{6}}}$. The choice of local coordinates is free, however, a natural choice is represented by exponential coordinates, therefore we define
\begin{equation}
\label{eq:local_coords_bas}
    \varepsilon = \vartheta\left(e\right) = \vartheta\left(e_{R},\,e_{b},\, e_{C}\right) = \left(\log\left(e_{R}\right)^{\vee},\, e_{b},\, \log\left(e_{C}\right)^{\vee}\right) \in \R^{9} ,
\end{equation}
with ${\vartheta\left(\xi_0\right) = \mathbf{0} \in \R^{9}}$. Let ${\Vector{}{u}{0} = \left(\Vector{}{\omega}{0},\,\mathbf{0},\,\mathbf{0},\,\mathbf{0}\right) \coloneqq \psi\left(\hat{X}^{-1}, \Vector{}{u}{}\right)}$ be the origin input, with ${\Vector{}{\omega}{0} \coloneqq \hat{A}\Vector{}{\omega}{} + \hat{a}^{\vee}}$.

The matrix ${\mathbf{A}_{t}^{0}}$ is defined by the following chain of differentials (Equ. 11)
\begin{equation*}
    \mathbf{A}_{t}^{0} = \Fr{e}{\xi_0}\vartheta\left(e\right)\Fr{E}{I}\phi_{\xi_0}\left(E\right)\Fr{e}{\xi_0}\Lambda\left(e, \Vector{}{u}{0}\right)\Fr{\varepsilon}{\mathbf{0}}\vartheta^{-1}\left(\varepsilon\right) .\\
\end{equation*}
Let us brake down the above expression and compute the matrix ${\mathbf{A}_{t}^{0}}$. Given the choice of local coordinates, the inverse of the chart transition map writes
\begin{equation*}
    e = \vartheta\left(\varepsilon\right)^{-1} = \vartheta\left(\varepsilon_{R},\,\varepsilon_{b},\,\varepsilon_{C}\right)^{-1} = \left(\exp\left(\varepsilon_{R}^{\wedge}\right),\, \varepsilon_{b},\, \exp\left(\varepsilon_{C}^{\wedge}\right)\right).
\end{equation*}
Therefore, from the Taylor expansion of the exponential we can write, ${\Fr{\varepsilon}{\mathbf{0}}\vartheta^{-1}\left(\varepsilon\right) = \eye}$. The second differential can be computed by differentiating the Lift ${\Lambda\left(e, \Vector{}{u}{0}\right)}$ with respect to $e$ and evaluating the differential at ${\xi_0 = \text{id}}$ as follows. Let $h \in T_{\xi_0}\calM$, then
\begin{equation*}
    \Fr{e}{\xi_0}\Lambda\left(e, \Vector{}{u}{0}\right)\left[h\right] = \lim_{t \to 0} \frac{1}{t}\left[\Lambda\left(\xi_0+ht, \Vector{}{u}{0}\right) - \Lambda\left(\xi_0, \Vector{}{u}{0}\right)\right] .
\end{equation*}
Computing and valuating the above expression at ${h = \Fr{\varepsilon}{\mathbf{0}}\vartheta^{-1}\left(\varepsilon\right)\left[\varepsilon\right] = \varepsilon}$ yields
\begin{equation*}
    \Fr{e}{\xi_0}\Lambda\left(e, \Vector{}{u}{0}\right)\left[\varepsilon\right] = \left(
    \begin{bmatrix}
        -\varepsilon_b^{\wedge} & -\Vector{}{\omega}{0}^{\wedge}\varepsilon_b\\
        \mathbf{0} & 0
    \end{bmatrix},\, \left(\Vector{}{\omega}{0}^{\wedge}\varepsilon_C\right)^{\wedge} - \varepsilon_b^{\wedge}
    \right).
\end{equation*}
The nex differential is computed differentiating ${\phi_{\xi_0}\left(E\right)}$ with respect to $E$ at identity. Let $h \in \gothg$, then
\begin{equation*}
    \Fr{E}{I}\phi_{\xi_0}\left(E\right)\left[h\right] = \lim_{t \to 0} \frac{1}{t}\left[\phi_{\xi_0}\left(I+ht\right) - \phi_{\xi_0}\left(I\right)\right] .
\end{equation*}
Computing and valuating the above expression at ${h = \Fr{e}{\xi_0}\Lambda\left(e, \Vector{}{u}{0}\right)\left[\varepsilon\right]}$ yields
\begin{equation*}
    \Fr{E}{I}\phi_{\xi_0}\left(E\right)\Fr{e}{\xi_0}\Lambda\left(e, \Vector{}{u}{0}\right)\left[\varepsilon\right] = \left(
    \begin{bmatrix}
        -\varepsilon_b^{\wedge} & \Vector{}{\omega}{0}^{\wedge}\varepsilon_b\\
        \mathbf{0} & 0
    \end{bmatrix},\, \left(\Vector{}{\omega}{0}^{\wedge}\varepsilon_C\right)^{\wedge}
    \right).
\end{equation*}
Finally, the last differential results in just taking the ${\cdot^{\vee}}$ operator, thus
\begin{equation*}
    \mathbf{A}_{t}^{0}\varepsilon = \left(-\varepsilon_b,\, \Vector{}{\omega}{0}^{\wedge}\varepsilon_b,\, \Vector{}{\omega}{0}^{\wedge}\varepsilon_C
    \right) = \begin{bmatrix}
        \mathbf{0} & -\mathbf{I} & \mathbf{0}\\
        \mathbf{0} & \Vector{}{\bm{\omega}}{0}^{\wedge} & \mathbf{0}\\
        \mathbf{0} & \mathbf{0} & \Vector{}{\bm{\omega}}{0}^{\wedge}
    \end{bmatrix}\begin{bmatrix}
        \varepsilon_R\\
        \varepsilon_b\\
        \varepsilon_C
    \end{bmatrix}
\end{equation*}